\documentclass[sigconf, nonacm]{acmart}

\AtBeginDocument{%
  }

\usepackage{times}
\usepackage{latexsym}
\usepackage{longtable}
\usepackage{array}

\usepackage{amsmath}
\usepackage{graphicx}
\usepackage{subcaption}
\usepackage{booktabs}
\usepackage{enumitem}

\begin{document}

\title{Robust Batch-Level Query Routing for Large Language Models under Cost and Capacity Constraints}



\author{Jelena Markovic-Voronov}
\affiliation{%
  \institution{LinkedIn}
  \country{}   
}

\author{Kayhan Behdin}
\affiliation{%
  \institution{LinkedIn}
  \country{} 
}

\author{Yuanda Xu}
\affiliation{%
  \institution{LinkedIn}
  \country{} 
}

\author{Zhengze Zhou}
\affiliation{%
  \institution{LinkedIn}
  \country{} 
}

\author{Zhipeng Wang}
\affiliation{%
  \institution{LinkedIn}
  \country{}
}
\authornote{Email: \texttt{zhipwang@linkedin.com}}

\author{Rahul Mazumder}
\affiliation{%
  \institution{LinkedIn, MIT}
  \country{}
}

\renewcommand{\shortauthors}{Markovic-Voronov et al.}

\begin{abstract}
  We study the problem of routing queries to large language models (LLMs) under cost, GPU resources, and concurrency constraints. Prior per-query routing methods often fail to control batch-level cost, especially under non-uniform or adversarial batching. To address this, we propose a batch-level, resource-aware routing framework that jointly optimizes model assignment for each batch while respecting cost and model capacity limits. We further introduce a robust variant that accounts for uncertainty in predicted LLM performance, along with an offline instance allocation procedure that balances quality and throughput across multiple models. Experiments on two multi-task LLM benchmarks show that robustness improves accuracy by 1–14\% over non-robust counterparts (depending on the performance estimator), batch-level routing outperforms per-query methods by up to 24\% under adversarial batching, and optimized instance allocation yields additional gains of up to 3\% compared to a non-optimized allocation, all while strictly controlling cost and GPU resource constraints.
\end{abstract}







\maketitle

\section{Introduction}

Large Language Models (LLMs,~\citep{llama3,achiam2023gpt,liu2024deepseek}) have become a central component of modern machine learning systems. Their strong performance is driven in part by substantial computational resources, particularly at inference time. However, allocating more inference resources does not always yield significantly better outputs, especially for simple queries~\citep{chen2024overthink}. This observation has motivated LLM routing~\citep{dinghybrid,nearestneigbor,matrixfactorization,song2025irt}, which aims to balance response quality and computational cost by routing easy queries to cheaper models and hard queries to more capable ones. LLM routing has become a major component of industrial agentic applications~\citep{withmartian_routing_for_ai_agents,openai2025gpt5systemcard,semanticrouter2025}.

Most existing routing methods operate on a per-query basis. Given a query $q$ and a pool of $M$ models $m_1,\ldots,m_M$, per-query routing methods estimate a quality score of how well model $m_j$ will work for query $q$, denoted as $l(q,m_j)$ and the cost of responding to query $q$ using model $m_j$, denoted by $c(q,m_j)$. Then, the routing is performed by solving
\begin{equation}\label{eq:perquery}
\max_{j\in[M]} \; l(q,m_j) - \lambda \cdot c(q,m_j),
\end{equation}
where the hyperparameter $\lambda\geq 0$ controls the cost–quality tradeoff. Prior work largely focuses on improving the estimator $l(q,m_j)$ using techniques such as neural networks, matrix factorization, k-nearest neighbors (kNN), or lightweight LLMs~\citep{mei2025omnirouter,nearestneigbor,matrixfactorization,song2025irt,feng2025graphrouter,zhuangembedllm}. 

In practice, however, LLM inference systems rely heavily on (dynamic) batching to improve hardware utilization, which inherently couples the cost, latency, and capacity usage of queries that arrive close in time~\citep{batch1,batch2}. This exposes several limitations of per-query routing in~\eqref{eq:perquery}. First, cost is only indirectly controlled through $\lambda$, making it difficult to enforce strict batch-level budget constraints. Second, per-query routing ignores batch-level effects: when multiple difficult queries arrive together, it may oversubscribe expensive or capacity-limited models, causing cost spikes or delays. Third, per-query routing does not account for heterogeneous serving setups, such as locally deployed models with limited capacity versus cloud-based models with higher cost but greater scalability~\citep{xin2025hybrid,pan2025cost}. Such serving settings have become increasingly common in industrial applications; For example, Apple~\citep{gunter2024apple} adopted a two-model approach, with a smaller model served on-device and a larger model served in the cloud. Finally, performance estimates $l(q,m_j)$ can be noisy at test time, leading to over- or under-confident routing decisions. We refer to Section~\ref{section:motivating} for an example demonstrating the shortcomings of per-query routing discussed here.

\begin{figure*}[htbp]
    \centering
    \includegraphics[width=0.9\textwidth]{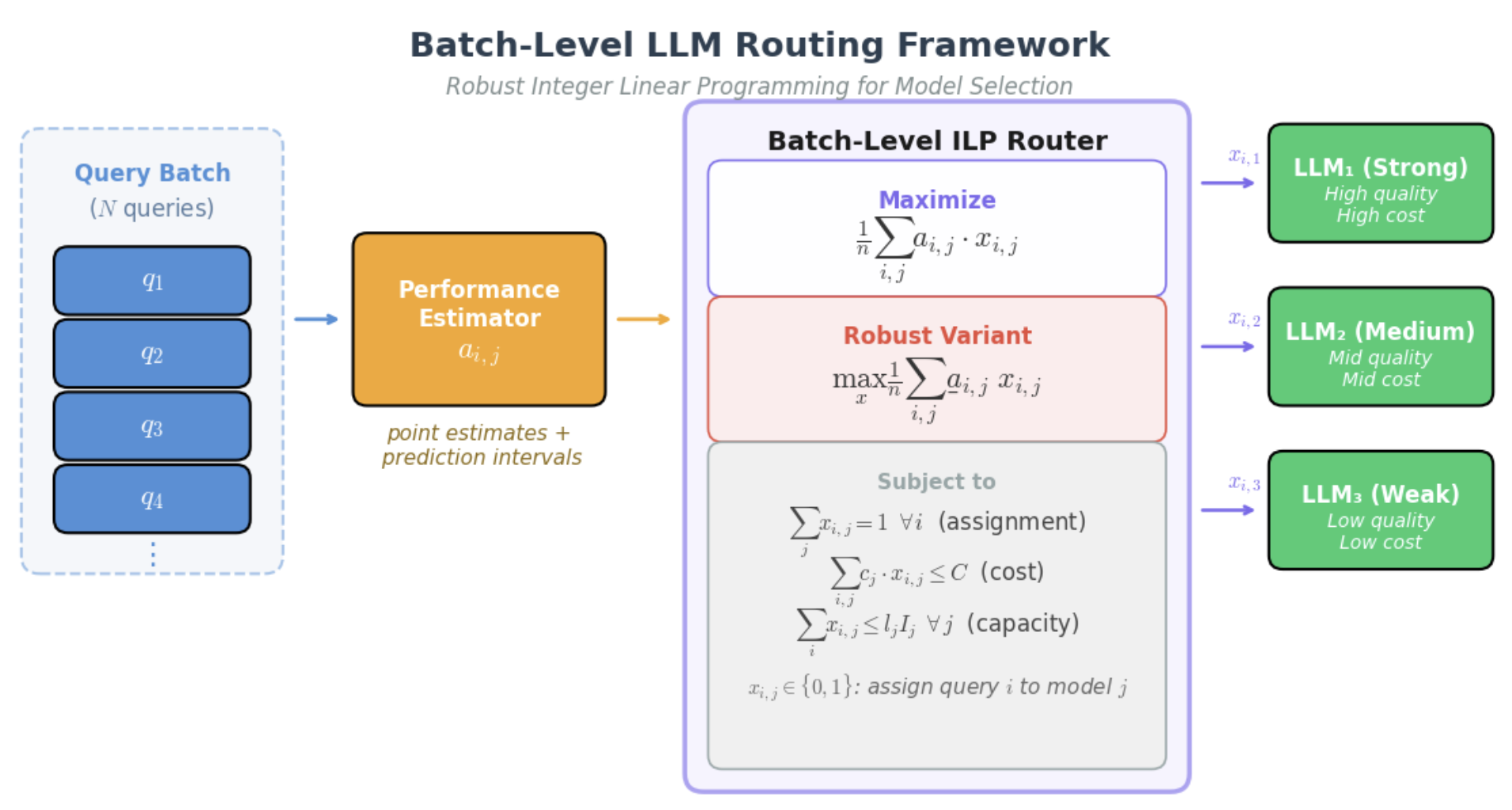}
\caption{Batch-level LLM routing framework assigns an LLM to each query within a batch by solving a constrained optimization problem that maximizes average per-query performance. The formulation enforces both a global monetary cost budget on total inference cost and individual capacity constraints for each LLM. Each query is assigned to exactly one LLM, so among $x_{i,1}, \ldots, x_{i,M}$ only one equals 1 and the rest are 0.
Since the true performance of an LLM on a new query is unknown at decision time, it must be estimated via $a_{i,j}$. To account for estimation uncertainty in $a_{i,j}$, the robust variant replaces the point performance estimates in the objective with the lower bounds $\underline{a}_{i,j}$ of their corresponding prediction intervals, thereby optimizing for worst-case performance within the estimated uncertainty range.}
\label{fig:diagram}
\end{figure*}

To address these issues, we propose a \textbf{batch-level routing framework} based on Integer Linear Programming (ILP,~\cite{nemhauser}), as illustrated in Figure \ref{fig:diagram}. Our formulation is designed to maximize the average routing quality across queries while explicitly enforcing both cost and hardware capacity constraints, which are critical in practical deployment scenarios. To account for uncertainty in the estimated performance $l(q,m_j)$ of model $m_j$ on query $q$, we introduce a robust optimization~\citep{ben2009robust} variant that ensures strong performance even under worst-case estimation errors. Importantly, the framework is model-agnostic and compatible with standard performance estimators such as XGBoost or k-nearest neighbors, and we demonstrate that the resulting ILPs can be solved efficiently using existing off-the-shelf solvers, enabling practical deployment for large-scale query batches.

Beyond inference-time routing, we further investigate optimal allocation of on-premise computational resources (e.g., GPUs) to available models prior to inference. This pre-deployment decision involves balancing accuracy and capacity across multiple models, ensuring that the system can handle incoming query batches without exceeding resource constraints. Unlike prior work on LLM routing, which typically assumes unlimited capacity or focuses solely on per-query performance, our framework explicitly integrates deployment planning, bridging the gap between offline resource allocation and online batch-level routing to achieve both efficient and high-quality inference.

\paragraph{Contributions}

We summarize the key contributions of this work as follows:
\begin{enumerate}[label=(\roman*)]
    \item We identify fundamental shortcomings of existing per-query LLM routing, particularly under batched inference and strict cost or capacity constraints.
    \item We introduce a robust batch-level routing framework that explicitly enforces cost and hardware constraints while accounting for uncertainty in performance estimates.
    \item We study the optimal allocation of local computational resources to models prior to inference.
    \item Extensive experiments on two multi-task LLM benchmarks show that our approach achieves more stable and higher routing accuracy than per-query methods while satisfying all constraints. Specifically, robustness improves accuracy by 1--14\% over non-robust counterparts, batch-level routing outperforms per-query methods by up to 24\% under adversarial batching, and optimized model instance allocation yields gains of up to 3\% compared to non-optimized model allocation, all while strictly controlling cost and GPU resources.
\end{enumerate}

\subsection{Related Work}

Most existing LLM routing methods follow a two-stage framework in which they first estimate the performance of each LLM on a given query and then optimize a per-query objective, as in~\eqref{eq:perquery}, to balance expected performance against cost. The primary distinction across prior work therefore lies in how LLM performance is modeled. For example, RouterBench~\citep{nearestneigbor} estimates per-query performance using either k-nearest neighbors or a multilayer perceptron (MLP), with a separate model trained for each LLM. UniRoute~\citep{dynamic} extends this approach by incorporating k-means clustering to enhance generalization to previously unseen LLMs. IRT-Router~\citep{song2025irt} employs a lightweight neural network that jointly encodes query features and LLM profile embeddings to produce performance estimates.  GraphRouter~\citep{feng2025graphrouter} models relationships between queries and LLMs through a graph neural network, leveraging structural information to improve prediction accuracy.

While most prior work assumes a fixed cost per LLM, some approaches incorporate query-dependent costs. 
GraphRouter~\citep{feng2025graphrouter} models query-dependent cost using the same graph neural network employed for performance estimation, whereas CARROT~\citep{somerstep2025carrot} combines kNN and transformer-based models to predict query-specific costs.

While the majority of the LLM routing methods consider the per-query objective in~\eqref{eq:perquery}, a few methods deviate. 
RouteLLM~\citep{matrixfactorization} restricts routing to two models and parameterizes the probability that the larger model outperforms the smaller one, avoiding explicit estimation of individual LLM performance and not optimizing an objective linear in cost. 
FORC~\citep{ilpnonbatched}, OmniRouter~\citep{mei2025omnirouter} and ~\citep{lpdual} solve a joint optimization problem across queries, making their approach closer in spirit to ours. We note that however neither consider batch-level routing, specially under resources constraints, as we do. We discuss these methods further in Section~\ref{subsection:opt:formulation}.

\section{Resource-Aware Model Selection}

\subsection{A Motivating Example} \label{section:motivating}

\begin{figure}
    \centering
    \includegraphics[width=\linewidth]{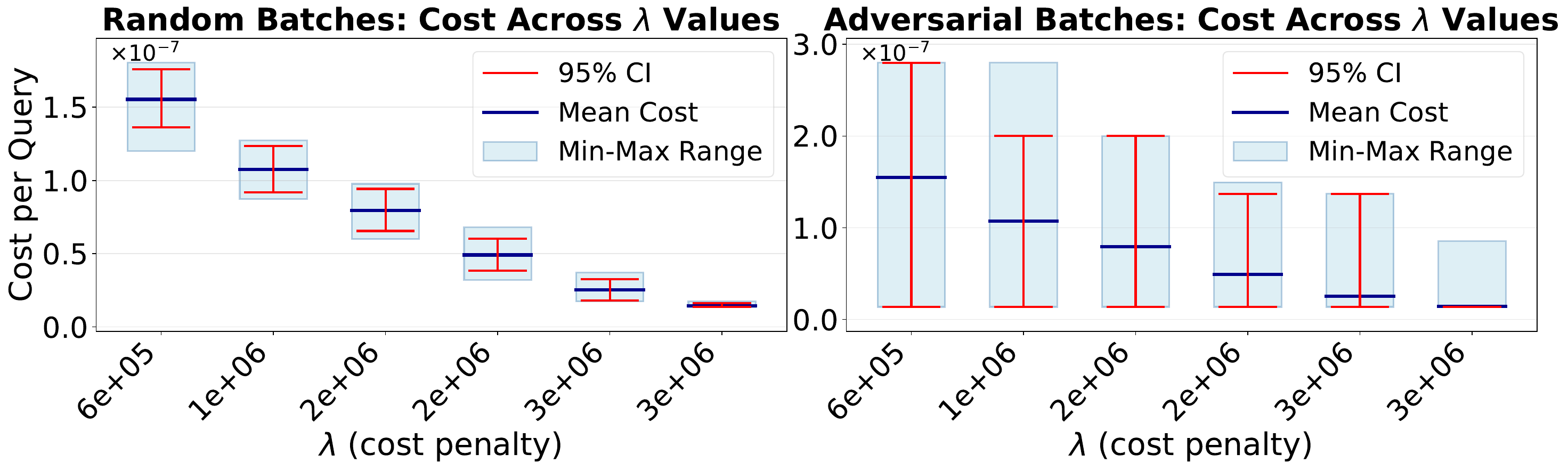}
    \caption{Comparison of cost inference of each batch for per-query routing~\eqref{eq:perquery}, for different values of $\lambda$. Routing is performed using MIRT, with batches are constructed either randomly to yield approximately uniform difficulty (left) or adversarially to create difficult and easy batches (right).}
    \label{fig:batch:cost}
\end{figure}

We demonstrate that per-query routing~\eqref{eq:perquery} can perform poorly under batched inference. Using the benchmark dataset and MIRT router of~\citep{song2025irt}, we apply the routing rule~\eqref{eq:perquery} with fixed values of $\lambda$ and report the average routing cost per query per batch of size 100 in Figure~\ref{fig:batch:cost}. We consider two settings: (i) batches formed uniformly at random, and (ii) batches constructed by grouping difficult queries together, resulting in non-uniform batch difficulty. Even with uniformly formed batches, per-batch inference cost varies substantially across different batches, for example, for $\lambda=6\times10^5$. The difference is even larger when the batches are non-uniform. This shows that it can be difficult to explicitly control per-batch inference cost using~\eqref{eq:perquery}, leading to periods that either the inference budget is not fully utilized, or the system goes over the budget.

\subsection{Online LLM Routing Optimization Formulation} \label{subsection:opt:formulation}

The router aims to maximize overall LLM performance subject to resource constraints, including inference cost, GPU availability, and latency-induced capacity limits. Consider a batch of $N$ queries and $M$ available models. For each model $j=1, \ldots, M$, we specify: (i) a per-query cost $c_j$, assumed to be query-independent~\citep{song2025irt} (though this assumption is not required in general); (ii) the number of available LLM model instances $I_j$, which is fixed due to the overhead of initializing LLMs on GPUs; and (iii) the per-model instance capacity $l_j$, defined as the maximum number of concurrent queries that can be routed to the model. For online batched routing, we treat $I_j$ as fixed; however, in Section~\ref{subsection:offline:model:allocation}, we compute the optimal $I_j$ in a data-dependent manner via a separate offline optimization.
The optimization parameters differ depending on the deployment setting: locally hosted models are typically constrained by hardware capacity (the number of GPUs), whereas cloud-based models are primarily constrained by (monetary) inference cost.

For each query–model pair $(i,j)$, let $a_{i,j}$ represent the estimated performance of model $j$ on query $i$, and let $x_{i,j} \in \{0,1\}$ indicate whether query $i$ is assigned to model $j$. Section~\ref{subsection:performance:estimates} discusses various methods to obtain the performance estimates $a_{i,j}$.
Given an average per-query budget $C$, we formulate the batch-level routing problem as:
\begin{align} \label{eq:routing}
\begin{aligned}
&\max_{\mathbf{x}\in \{0,1\}^{N\times M}} ~~
    \frac{1}{N} \sum_{i=1}^N \sum_{j=1}^M a_{i,j} \, x_{i,j} & \text{(performance)} \\
&~\text{s.t.} \quad \frac{1}{N}\sum_{i=1}^N \sum_{j=1}^M c_j \, x_{i,j} \le C, &  \text{(cost)} \\
& \quad \quad \sum_{i=1}^N x_{i,j} \le l_j I_j \quad \forall j\in[M], &  \text{(capacity)} \\
& \quad \quad \sum_{j=1}^M x_{i,j} = 1 \quad \forall i \in[N]. &  \text{(assignment)} \\
\end{aligned}
\end{align}
where $[M]$ denotes the index set $\{1, \ldots, M\}$.
The objective maximizes the average routing quality over the batch, while the constraints ensure that the total batch cost does not exceed $N\cdot C$ and that each model’s capacity, $l_j I_j$, is not violated. We note that this optimization must be solved for every batch of concurrent queries, potentially multiple times per second.

Problem~\eqref{eq:routing} is a (non-convex) integer linear program (ILP). While ILPs are generally challenging to solve due to non-convexity, recent advances in integer programming have made them tractable at small to moderate scales. In particular, off-the-shelf solvers such as SCIP~\citep{SCIPOptSuite10} can solve such ILPs in milliseconds. As we show empirically, solving the per-batch routing problem in~\eqref{eq:routing} introduces no significant runtime overhead.

\paragraph{Related Work}
Although Problem~\eqref{eq:routing} is novel, it is closely related to prior optimization-based LLM routing approaches~\citep{lpdual,ilpnonbatched,mei2025omnirouter}. 
\cite{ilpnonbatched} formulate routing as an integer linear program (ILP), but assume that all queries are known in advance, instead of arriving in batches. This is an unrealistic assumption in online settings, and in addition, they do not incorporate capacity constraints needed to distinguish between local and cloud deployments. 
Similarly, \cite{mei2025omnirouter} propose a convex continuous relaxation of the original integer program, but do not incorporate capacity constraints. Importantly, solving the relaxed convex formulation can yield solutions that differ substantially from those of the original discrete optimization problem.
While \cite{lpdual} also consider an ILP-based formulation, their method ultimately reduces to per-query routing, thereby inheriting its limitations, and likewise does not model differences between local and cloud capacity.


\subsection{Performance Estimates} \label{subsection:performance:estimates}

The effectiveness of the router critically depends on the accuracy of the estimated model performance, i.e., how closely the predicted values $a_{i,j}$ match the true performance of model $j$ on query $i$. We however note that our batch-level routing framework is compatible with any performance estimation method. Following prior work, we consider kNN~\citep{nearestneigbor} and multidimensional item response theory (MIRT)-based estimators~\citep{song2025irt}, which leverage neural networks to predict LLM performance. The use of MIRT is motivated by recent findings showing that MIRT-based routers outperform competing approaches~\citep{lu2025routerarena}.

In addition, we propose an estimator based on XGBoost. XGBoost is a scalable gradient-boosted decision tree framework~\citep{chen2016xgboost} that is well-suited for this setting due to its strong empirical performance on tabular data, fast training and inference, and ability to capture non-linear interactions without extensive feature engineering. These properties make XGBoost particularly attractive for routing systems, where performance estimates must be computed efficiently and updated frequently.

\subsection{Robust LLM Routing}

Regardless of the method used to estimate the values $a_{i,j}$ in~\eqref{eq:routing}, these estimates inevitably involve some degree of uncertainty. For instance, if certain $a_{i,j}$ values are overestimated, some queries may be routed to models that underperform. To address this, we introduce a robust routing formulation that maximizes routing quality under a worst-case scenario with respect to the estimated $a_{i,j}$ values. Such robust routing is particularly desirable in practice, where the system must meet a minimum quality threshold even in adverse conditions.
Specifically, we assume that for each pair $(i,j)$, a prediction interval $[\underline{a}_{i,j}, \widebar{a}_{i,j}]$ can be estimated such that, with high probability, $a_{i,j}$ lies within $[\underline{a}_{i,j}, \widebar{a}_{i,j}]$. To make~\eqref{eq:routing} robust, we replace the optimization objective with the worst-case performance 
\begin{equation}\label{eqn:obg-robust}
\min_{a_{i,j}\in[\underline{a}_{i,j},\widebar{a}_{i,j}]} \frac{1}{N} \sum_{i=1}^N \sum_{j=1}^M a_{i,j} \, x_{i,j}.
\end{equation}
Given a set of query-model assignment $x_{i,j}$'s,~\eqref{eqn:obg-robust} quantifies the average quality, under a worst-case scenario for all $a_{i,j}$'s (this is the worst possible accuracy that we might observe for this batch). Such formulations are commonly referred to as being robust~\citep{ben2009robust}. By maximizing this worst-case accuracy, we ensure the routing does not catastrophically fail. Such guarantees might be interesting in production and real-world deployments, where even under adverse conditions, the model should perform well.  

As $x_{i,j}\in\{0,1\}$, it is straightforward to see that using the robust objective~\eqref{eqn:obg-robust} is equivalent to replacing the original objective in~\eqref{eq:routing} with 
$$\frac{1}{N} \sum_{i=1}^N \sum_{j=1}^M \underline{a}_{i,j} \, x_{i,j}.$$
In other words, instead of using a point estimate for $a_{i,j}$, robust routing relies on the lower bound $\underline{a}_{i,j}$ of its prediction interval. This is intuitive: to improve worst-case performance, we should avoid overestimating model quality. We note that this robust formulation can also be applied at the per-query routing level.

Regarding the estimation of the lower bound $\underline{a}_{i,j}$, this corresponds to constructing a statistically valid uncertainty estimate for the predicted performance. This is a well-studied problem in statistics. A standard approach is bootstrap resampling~\citep{bootstrap}, where multiple datasets are generated by sampling with replacement from the training data, and the performance estimator is refit on each resample. This yields a distribution of predicted values for $a_{i,j}$, from which one can derive a lower prediction bound (e.g., via an empirical quantile).
When repeated refitting is computationally expensive or otherwise impractical, conformal prediction~\citep{conformal} provides an attractive alternative.

\subsection{Offline Model Allocation Formulation} \label{subsection:offline:model:allocation}

In Problem~\eqref{eq:routing}, we assume that the total number of available instances for each model, $I_j$, is given. For example, local LLMs may be served by a fixed number of on-premise GPUs. A key question is how to allocate these resources across different model choices. One option is to deploy many small models (larger $I_j$ for smaller models), increasing the total local serving capacity, but potentially sacrificing quality. Conversely, one could allocate GPUs to a few large models to maintain high quality, but this limits local throughput and increases overall inference cost. To determine an optimal assignment of $I_j$, we propose an optimization formulation that is solved prior to inference. Once the optimal $I_j$ values are set, inference proceeds with these allocations fixed.

Our goal is to simulate the (online) inference-time performance of the router during the (offline) resource assignment phase. We assume we have access to $B$ held-out calibration batches of the routing data (that were not used to train the $a_{i,j}$ estimators). Specifically, we let $a_{i,j}^b \in [0,1]$ denote an estimated performance score measuring how well model $j$ answers query $i$ in batch $b$ (or the lower prediction interval, in case of a robust formulation). For each model $j$, we deploy $I_j$ instances each consuming $g_j$ GPUs, where $g_j$ is given a priori. To simulate the system's inference-time performance, we aim to maximize average performance, while respecting inference cost, GPU constraints and concurrency constraints over $B$ batches:
\begin{align}\label{eq:assignment}
\begin{aligned}
    &\max_{\mathbf{x}\in \{0,1\}^{N\times M}, \:\mathbf{I}\in \mathbb{Z}_{\geq 0}^M}  \quad \frac{1}{N\cdot B}
    \sum_{b=1}^B \sum_{i=1}^N \sum_{j=1}^M a_{i,j}^b \, x_{i,j}^b &  \text{(performance)} \\ 
    &\text{ s.t.} \quad \frac{1}{N\cdot B} \sum_{b=1}^B \sum_{i=1}^N \sum_{j=1}^M c_j \, x_{i,j}^b
    \le C, &  \text{(cost)}\\
    &\quad \quad \sum_{j=1}^M g_j I_j \le G, &  \text{(model allocation)}\\
    & \quad \quad \sum_{i=1}^N x_{i,j}^b \le l_j I_j \quad \forall j \in[M]\quad \forall b \in[B], &  \text{(capacity)}\\
    &\quad\quad  \sum_{j=1}^M x_{i,j}^b = 1 \quad \forall i\in[N] \quad \forall b\in[B]. &  \text{(assignment)}\\
\end{aligned}
\end{align}
Problem~\eqref{eq:assignment} simultaneously optimizes the binary variables $x_{i,j}^b$ and the nonnegative integer variables $I_j$. After solving this problem offline and determining the values of $I_j$ for $j = 1, \ldots, M$, these quantities are fixed and subsequently used as inputs to the online batch-level optimization problem in~\eqref{eq:routing}.

\section{Experiments}

This section describes the experimental setup and results obtained in a simulation environment, which we use to evaluate the benefits of our modeling and optimization choices, including robustness (Section \ref{subsection:robust:opt:per:query}), batching (Section \ref{subsection:model:batch-level}), data-dependent selection of the number of model instances (Section \ref{subsection:model:instances}), and full routing optimization with cost and resource constraints (Section \ref{subsection:model:full}).

\paragraph{Datasets}
We conduct experiments on two public datasets that include queries, LLM responses, associated scores, and costs required for routing. 
\begin{enumerate}
\item \textbf{Dataset 1.} The in-distribution benchmark introduced by the state-of-the-art work of~\citet{song2025irt}. This combined benchmark aggregates eight tasks (MMLU, CMMLU, ACLUE, ARC\_C, Hotpot\_QA, SQUAD, MATH, MBPP) and is split by task into 70\% training and 30\% test sets, containing 24,414 and 10,467 queries, respectively. For each query--LLM pair across the 20 considered LLMs, the dataset provides LLM responses and corresponding performance scores in $[0,1]$. It also includes per-input-token and per-output-token costs for each LLM, along with a brief profile describing each model's primary capabilities. 


\item \textbf{Dataset 2.} The benchmark dataset introduced in~\citet{nearestneigbor}, which aggregates eight tasks (MMLU, Hellaswag, GSM8K, ARC Challenge, Winogrande, MBPP, MT-Bench) and is split by task into 70\% training and 30\% test sets, comprising 25,547 training queries and 10,950 test queries. For each query, the dataset contains responses and scores from 11 LLMs. It provides the cost for each query--LLM pair, and we compute the cost of each model as the average cost across queries. We construct LLM profiles for this dataset following the same profile structure used in Dataset 1.


\end{enumerate}

More details on LLM pricing and model profiles are provided in Appendix~\ref{subsection:llm:pricing:profiles}.

\subsection{Robust Estimation Performance} \label{subsection:robust:opt:per:query}

To analyze robust estimation, we compare several performance estimators within a standard per-query optimization framework as in~\eqref{eq:perquery}, varying the performance–cost trade-off parameter $\lambda$. Data-specific choices include:

\begin{enumerate}
    \item For Dataset 1, the cost in the optimization objective is based on the output-token cost for each LLM, while the reported total cost observed in simulations reflects the total cost over all test queries. This total cost is computed as the sum of the input cost (\$/input token $\times$ query length) and output cost (\$/output token $\times$ output length). This follows prior work's setup~\citep{song2025irt}. All methods use BERT embeddings~\citep{devlin2019bert} to represent both queries and LLM profiles.
    
    \item For Dataset 2, the cost of each LLM is computed using the average per-query cost of each LLM, and the total cost is obtained by summing costs across all queries. All methods ran on this dataset use all-MiniLM-L12-v2 embedding model from SentenceTransformers~\citep{reimers2019sentence} to represent both queries and LLM profiles. The cost and the embedding model are chosen following the same setup as reported in \cite{nearestneigbor}.
\end{enumerate}

For each dataset, we evaluate MIRT~\citep{song2025irt}, XGBoost, and kNN, along with robust variants of XGBoost and kNN. 

\paragraph{Methods Configuration and Robust Performance Estimation}

For XGBoost, the input consists of the concatenation of the query and LLM profile embeddings, and hyper-parameters are selected via cross-validation. We vary the number of neighbors in the kNN, considering values of $k \in \{5, 40\}$. The smaller value ($k=5$) captures highly local structure and emphasizes the most similar examples, while the larger value ($k=40$) enables broader neighborhood aggregation and smoother predictions. Notably, $k=40$ was reported as the optimal choice in~\citet{nearestneigbor}. To obtain robust predictions for XGBoost and kNN, we apply bootstrap resampling to capture performance variability. We generate 100 bootstrap samples of the training data, refit the model on each sample, and obtain 100 predictions per test query. These predictions form an empirical distribution, and the robust estimate is defined as its empirical $Q$\% quantile (the lower bound of a one-sided $1-Q$\% prediction interval). This yields a conservative performance estimate that accounts for estimation uncertainty.

\begin{figure}[htbp]
\centering

\begin{subfigure}{\columnwidth}
    \centering
    \includegraphics[width=\columnwidth]{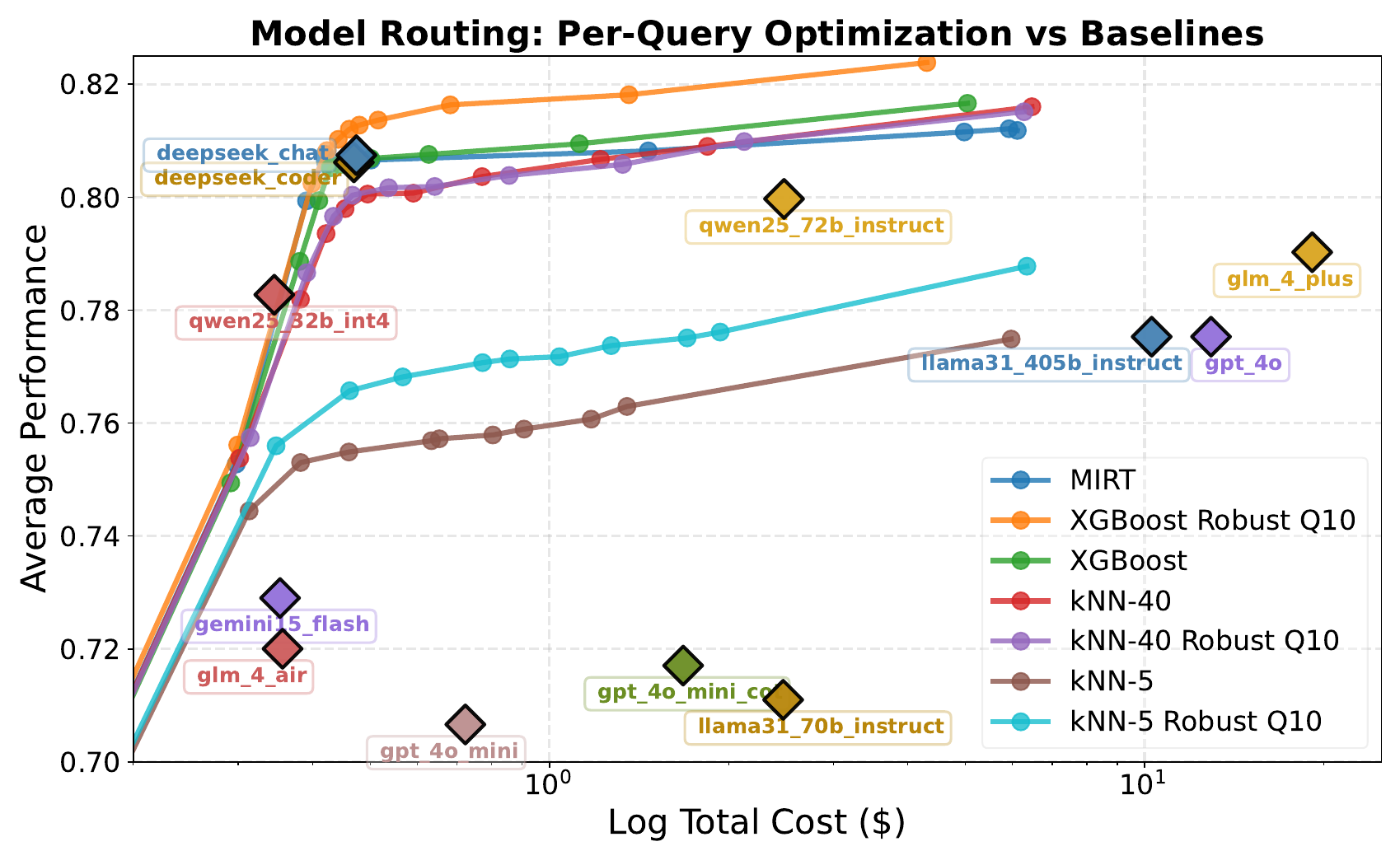}
    \caption{Dataset 1}
    \label{fig:per:query:opt:dataset:1}
\end{subfigure}

\vspace{0.5em} 

\begin{subfigure}{\columnwidth}
    \centering
    \includegraphics[width=\columnwidth]{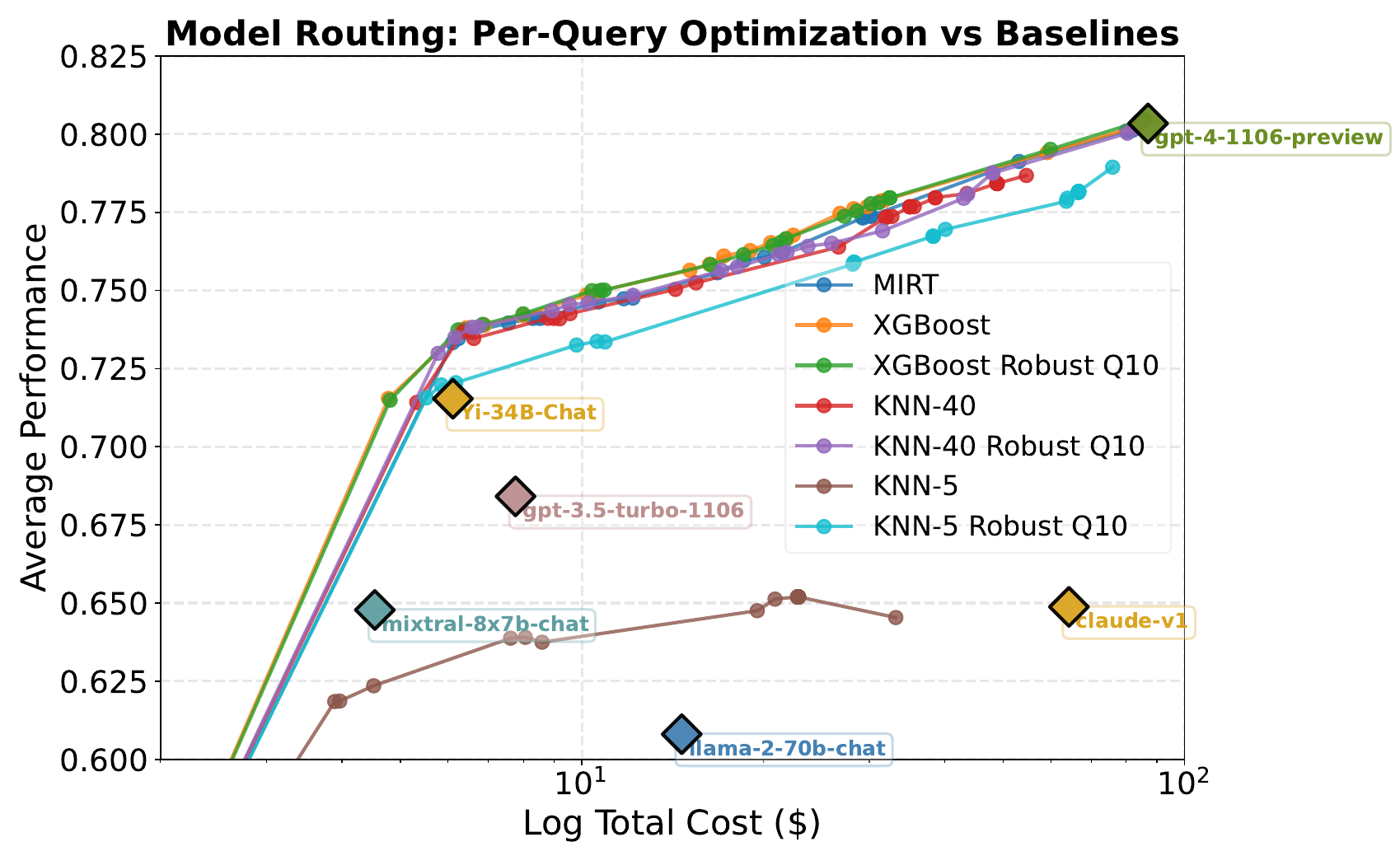}
    \caption{Dataset 2}
    \label{fig:per:query:opt:dataset:2}
\end{subfigure}

\caption{Average test set performance as a function of log total cost under per-query optimization for the two datasets. Each curve corresponds to a different performance estimator and shows the trade-off across varying $\lambda$ parameter. The robust estimators are calculated based on 10\% quantile ($Q=10$). We also include individual LLMs in the plots; these points correspond to benchmark-reported performance and cost values used as baselines in the simulation. For plot readability, we restrict these to models whose average test performance exceeds 70\% on Dataset 1 and 60\% on Dataset 2.}
\label{fig:per:query:opt}
\end{figure}

\paragraph{Per-query Routing Performance of Robust vs.~Non-Robust Methods}
\begin{enumerate}
    \item For Dataset 1, Figure~\ref{fig:per:query:opt:dataset:1} shows that, for any given cost, robust XGBoost achieves the highest test performance among all methods, outperforming MIRT, kNN-5, kNN-40, their robust versions, and non-robust XGBoost. kNN-40 performance is comparable to its robust version, while the robust version of kNN-5 significantly outperforms its non-robust counterpart.
    Moreover, robust XGBoost consistently outperforms any individual LLM at the same cost. Notably, the robust XGBoost estimator achieves up to 1.2\% accuracy improvement over the state-of-the-art MIRT router for certain cost values, and is the only routing method that outperforms the best individual model, DeepSeek\_Chat, at the same cost as DeepSeek\_Chat. 
    
    \item For Dataset 2, Figure~\ref{fig:per:query:opt:dataset:2} shows that routing based on XGBoost and its robust variant achieves the best performance -- with no significant difference between them -- while both outperform MIRT and kNN-based methods, including the robust versions of kNN.
    In contrast, robustness yields substantial gains for kNN-5 and kNN-40, improving performance by up to 1.7\% and 14.4\%, respectively.
\end{enumerate}

The results reported here for the robust methods use $Q=10$; in Appendix~\ref{subsection:per:query:quantiles}, we also evaluate these methods for $Q \in \{5, 10, 20, 30\}$, and find that the performance across these quantiles shows no significant differences.

Overall, the results indicate that incorporating robustness can improve the router’s performance in some cases, while in others it does not compromise performance, with routers based on both XGBoost and its robust variant consistently ranking among the top performers.

\paragraph{Dataset-Driven Differences in Routing Behavior}

Dataset 1 contains many competitive LLMs with similar (high) performance but different inference costs, resulting in relatively small performance gaps, whereas Dataset 2 features less competitive LLMs with substantially larger quality differences. 
Accordingly, per-query routing achieves meaningful cost–quality trade-offs on Dataset 1 and can outperform any individual LLM at the same cost level. In contrast, on Dataset 2, it cannot substantially improve over the strongest model, GPT-4. Its behavior instead approaches that of the Zero Router~\citep{nearestneigbor}, which linearly interpolates along the cost–quality frontier by randomly selecting among models that maximize expected quality under a fixed budget. Nevertheless, even in this regime, the learned router still provides consistent performance improvements over the Zero Router baseline for some costs.

\paragraph{Robust Estimation: Risk-Aware Performance}

To clearly separate the effects of robustness-aware optimization from standard (non-robust) selection, we perform a detailed per-query comparison of the resulting model choices. In this analysis, we omit cost considerations. For each query, we record (i) the LLM selected by the non-robust method, which chooses the LLM model with the best point estimate of predicted performance and does not account for uncertainty, and (ii) the LLM selected by our robust optimization procedure, which explicitly incorporates uncertainty estimates derived from bootstrap resampling into the selection objective. 

To quantify uncertainty for each selected LLM, we construct 95\% prediction intervals (PIs) using bootstrap resampling, following the same procedure as before.
We use the length of the 95\% PI as a proxy for predictive uncertainty and risk. A shorter interval indicates that the predicted performance is more concentrated across bootstrap samples, reflecting greater stability and lower uncertainty. In contrast, a longer interval suggests higher variability in the predictions, indicating greater uncertainty and risk.

Table~\ref{tab:pi_comparison} summarizes, across all queries, the percentage of cases in which the LLM chosen by the non-robust method has a shorter, equal, or longer PI than the LLM selected by the robust approach. The results show that, in the majority of queries, the non-robustly selected LLM exhibits a longer prediction interval than the robustly selected one. In other words, when uncertainty is ignored during model selection, the chosen LLM tends to have higher variability in its performance estimates.

These findings indicate that robustness-aware optimization systematically favors LLMs with lower predictive uncertainty, effectively reducing risk compared to point-estimate–based selection.

\begin{table}[htbp]
\centering
\begin{subtable}[t]{0.48\textwidth}
\centering
\begin{tabular}{lccc}
\hline
Method & < Rob.~PI & = Rob.~PI & > Rob.~PI  \\
\hline
XGBoost & 23.5 & 34.7 & 41.8 \\
kNN-40 & 5.3  & 83.7 & 11.0 \\
kNN-5  & 3.1  & 77.1 & 19.7 \\
\hline
\end{tabular}
\caption{Dataset 1}
\label{tab:pi_robust}
\end{subtable}

\vspace{0.5em} 

\begin{subtable}[t]{0.48\textwidth}
\centering
\begin{tabular}{lccc}
\hline
Method & < Rob.~PI & = Rob.~PI & > Rob.~PI \\
\hline
XGBoost & 0.5  & 98.2 & 1.2 \\
kNN-40 & 1.0  & 44.6 & 54.4 \\
kNN-5  & 3.2  & 36.9 & 59.9 \\
\hline
\end{tabular}
\caption{Dataset 2.}
\label{tab:pi_boot}
\end{subtable}

\caption{Comparison of prediction interval (PI) lengths for the LLMs selected by the main (non-robust) method versus the PI lengths of the LLMs selected by its robust version, shown as percentages of all queries. 
\(<\) Rob.~PI: percentage of queries where the non-robust PI is shorter than the robust PI; 
\(=\) Rob.~PI: percentage where they are equal; 
\(>\) Rob.~PI: percentage where the non-robust PI is longer than the robust PI.}
\label{tab:pi_comparison}
\end{table}

\subsection{Batch-Level Optimization for Worst-Case Cost Control} \label{subsection:model:batch-level}

Continuing the example from Section~\ref{section:motivating}, we compare batch-level optimization with per-query optimization under both random and adversarial batching schemes. Batch-level optimization~\eqref{eq:routing} directly enforces a cost constraint for each batch, while per-query optimization~\eqref{eq:routing} requires careful tuning of $\lambda$ to indirectly control the worst-case batch cost.

We partition the test set into random and adversarial batches of size $B$. Adversarial batches are constructed by sorting queries according to the cost of their selected LLMs and forming batches from consecutive sorted queries, thereby concentrating the highest-cost queries in some batches and the lowest-cost queries in others. For per-query optimization, we sweep over several values of the cost parameter $\lambda$ and compute, for each setting, the maximum batch-level average per query cost across batches. We then use this maximum observed average per query cost as the cost per query budget constraint $C$ in the batch-level optimization problem~\eqref{eq:routing}. For a fair comparison, we exclude GPU and capacity constraints in this section, and evaluate the resulting average performance across batches. In addition, based on experiments from Section~\ref{subsection:robust:opt:per:query}, we focus on the robust XGBoost $Q=10$ performance in this section, as it appears to be the most promising.

\begin{figure}[htbp]
\centering

\begin{subfigure}{\columnwidth}
    \centering
    \includegraphics[width=\columnwidth]{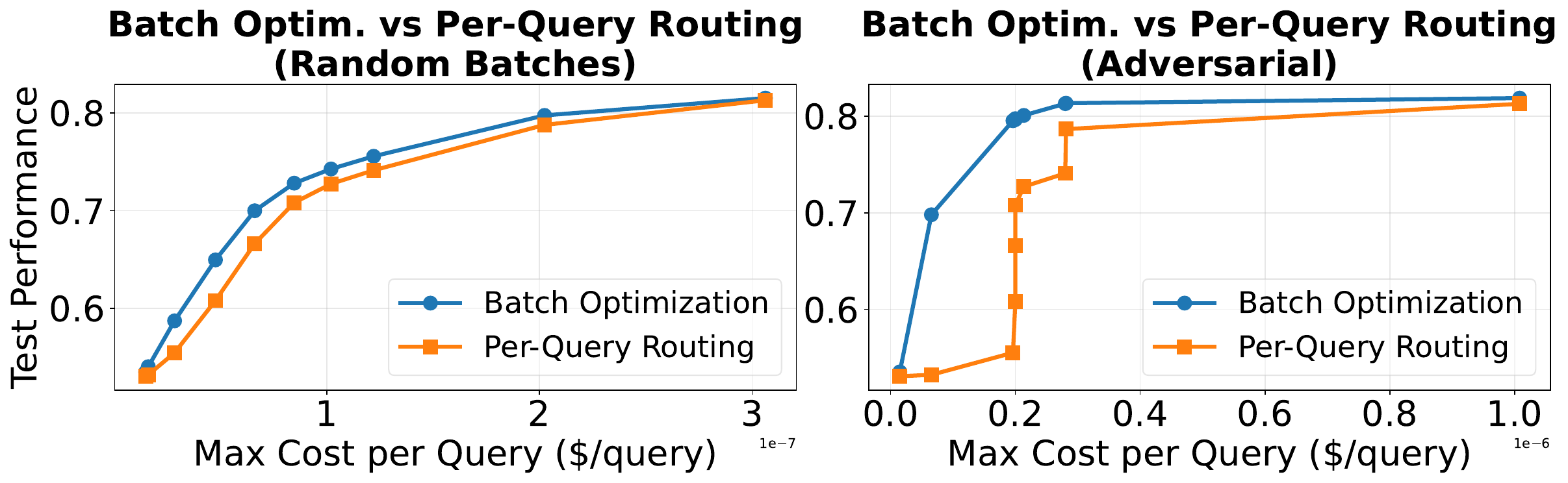}
    \caption{Dataset 1: $B=100$}
\end{subfigure}

\vspace{0.5em} 

\begin{subfigure}{\columnwidth}
    \centering
    \includegraphics[width=\columnwidth]{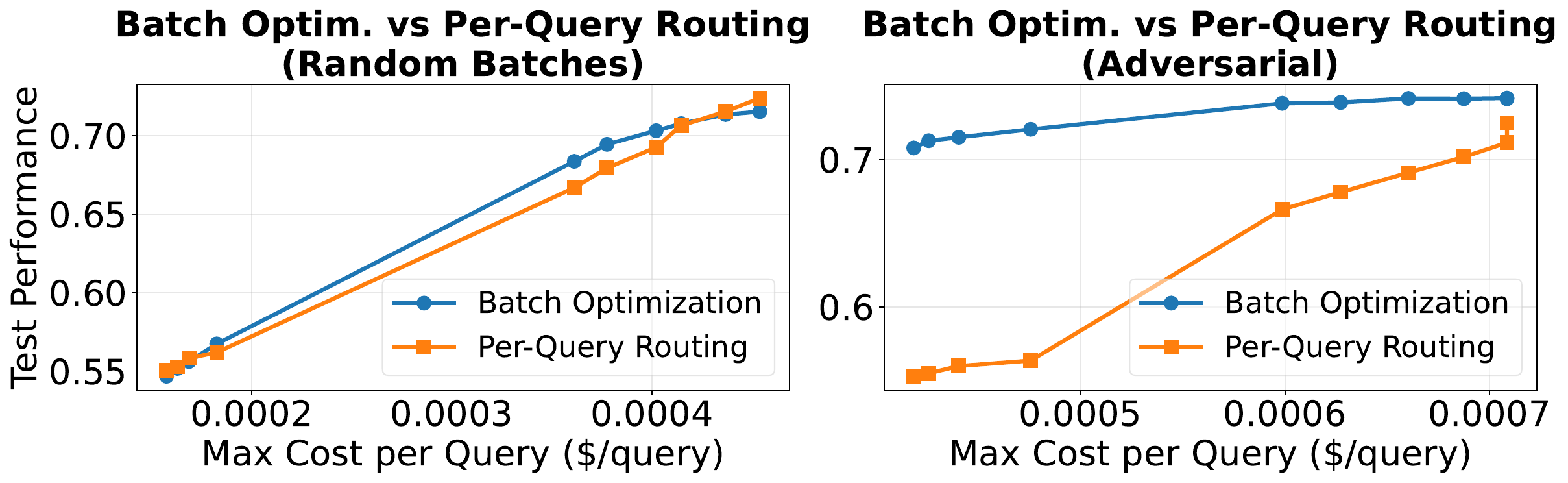}
    \caption{Dataset 2: $B=100$}
\end{subfigure}

\caption{Average test performance versus maximal batch-level average per-query cost across batches, comparing per-query optimization and batch-level optimization for both datasets. Results are shown for randomly constructed batches (left) and adversarial batches (right).}
\label{fig:batch:optimization:performance:cost}
\end{figure}

\paragraph{Batch-Level vs.~Per-Query Optimization Results}
Figure~\ref{fig:batch:optimization:performance:cost} reports results for batch size $B=100$ on both datasets under varying cost budgets. Across both random and adversarial batching settings, batch-level optimization consistently achieves higher test performance than per-query routing, while ensuring that the total cost of each batch remains below the prescribed maximum budget. This advantage holds uniformly across the range of cost constraints, indicating that explicitly optimizing at the batch level leads to more effective allocation of computational resources.

Overall, batch-level routing yields substantial gains over per-query routing, improving performance by up to:
\begin{enumerate}
    \item 4\% under random batching and 24\% under adversarial batching on Dataset 1, and
    \item 1.7\% under random batching and 15.8\% under adversarial batching on Dataset 2.
\end{enumerate}
These results suggest that batch-level routing better utilizes the available budget by reducing cost variance across batches and preventing inefficient budget allocation to individual queries. Additional experiments for other batch sizes $B\in\{50,200,400\}$ are provided in Appendix~\ref{subsection:cost:additional:exp}.


\subsection{Optimizing Model Instance Allocation}
\label{subsection:model:instances}

\paragraph{Deployment Setting: Costs and GPU Requirements for API-Based vs.~Self-Hosted LLMs}
We consider a deployment setting in which closed-source models are accessed via external APIs and incur per-inference monetary costs, without requiring any GPU allocation on our side. In contrast, open-source models are hosted internally and therefore require dedicated GPU resources but do not incur additional per-inference monetary costs. We assume that large open-source models with hundreds of billions of parameters require up to eight GPUs for initialization, while smaller models can be initialized with as few as one GPU. Table~\ref{table:llm:pricing:gpus} in Appendix~\ref{subsection:llm:pricing:gpus} gives the costs and GPU requirements for all models considered for both datasets. For ease of comparison, we set the concurrency parameter $l_j$ to 1 for all open-source models. This assumption is made solely for experimental consistency and can be easily adjusted to reflect different deployment configurations or hardware setups.

\paragraph{Fixed vs.\ Optimized Instance Allocation}

We compare average performance across batches under two settings: (i) a fixed allocation of open-source model instances, without offline optimization, and (ii) a data-dependent allocation obtained via optimization in~\eqref{eq:assignment}. In the fixed-allocation setting, we assign the same number of instances $n$ to each open-source model and vary $n$, so that $I_j=n$ for all $j=1,\ldots,M$.  In the optimized setting, we allow the optimizer to allocate the number of open-source model instances across models in a data-dependent manner using the offline optimization from \eqref{eq:assignment}, subject to the same total GPU budget constraint as the fixed allocation. We use 10\% of the test set for offline computation of the optimal number of model instances. Given the model instances -- whether fixed or optimized -- the remaining 90\% of the test data is used for the online evaluation of the batched optimization in~\eqref{eq:routing}. 
As in Section~\ref{subsection:model:batch-level}, we focus on the robust XGBoost $Q=10$ model fit on the training data. The batch size is set to $B=100$ here; additional experiments for different batch sizes $B\in \{50,200, 400\}$ are reported in Appendix~\ref{subsection:llm:instances:additional:exp}. 

\paragraph{SCIP Solver for Batch-Level Optimization}
We employ the SCIP solver~\citep{SCIPOptSuite10} to solve both the offline and batched online optimization problems. SCIP (Solving Constraint Integer Programs) is a state-of-the-art solver for mixed-integer and combinatorial optimization, combining branch-and-bound, cutting planes, and constraint propagation techniques to efficiently handle large and complex ILPs. In our experiments, SCIP requires under 0.4 seconds for larger batch sizes ($B=400$) and as little as 0.09 seconds for smaller batch sizes ($B=50$), demonstrating its suitability for real-time batch-level LLM routing.

\paragraph{Performance Gains from Optimized Allocation of Model Instances Under Different Cost Budgets}

\begin{figure}[htbp]
\centering

\begin{subfigure}{\columnwidth}
    \centering
    \includegraphics[width=\columnwidth]{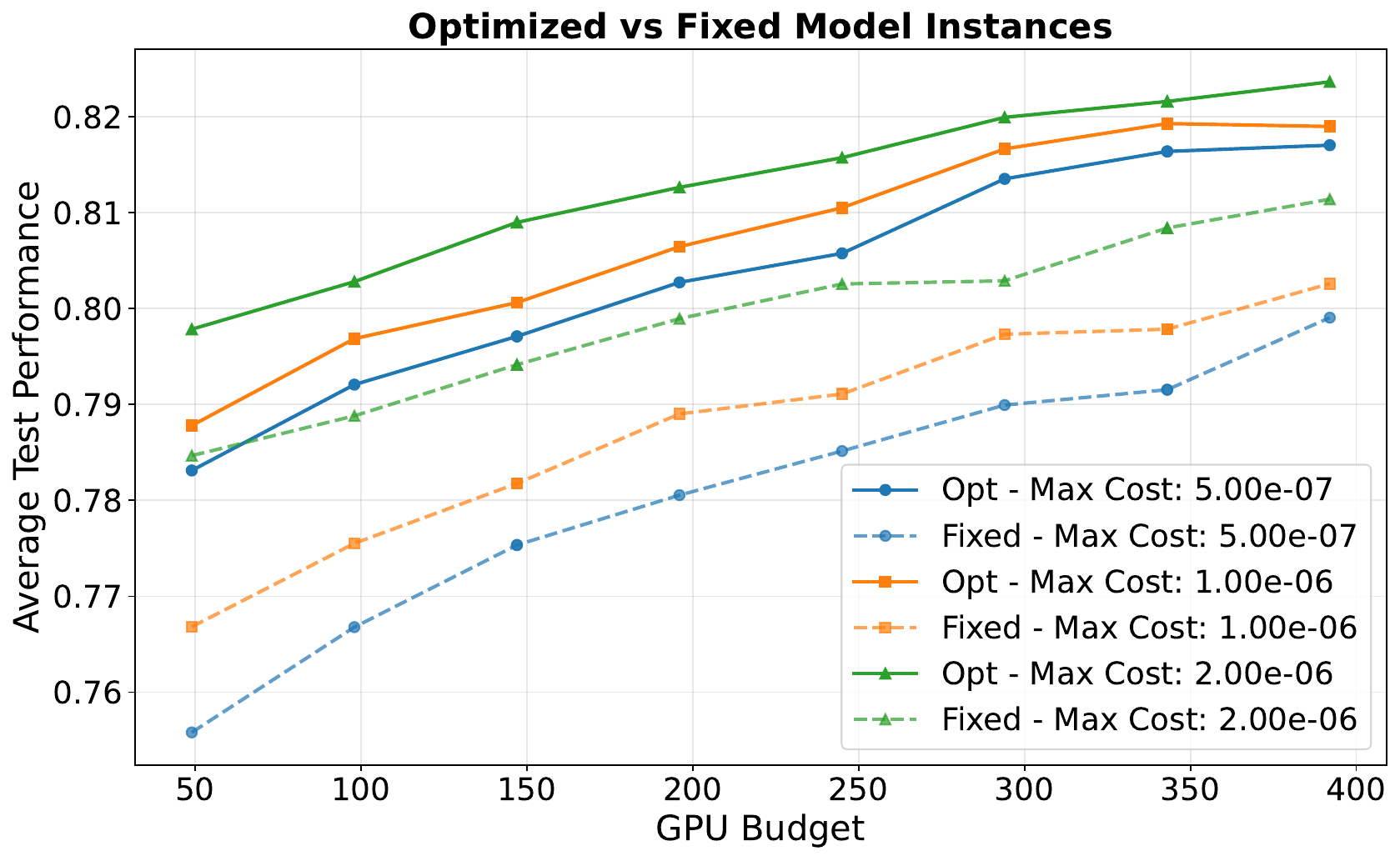}
    \caption{Dataset 1: $B=100$}
\end{subfigure}

\vspace{0.5em} 

\begin{subfigure}{\columnwidth}
    \centering
    \includegraphics[width=\columnwidth]{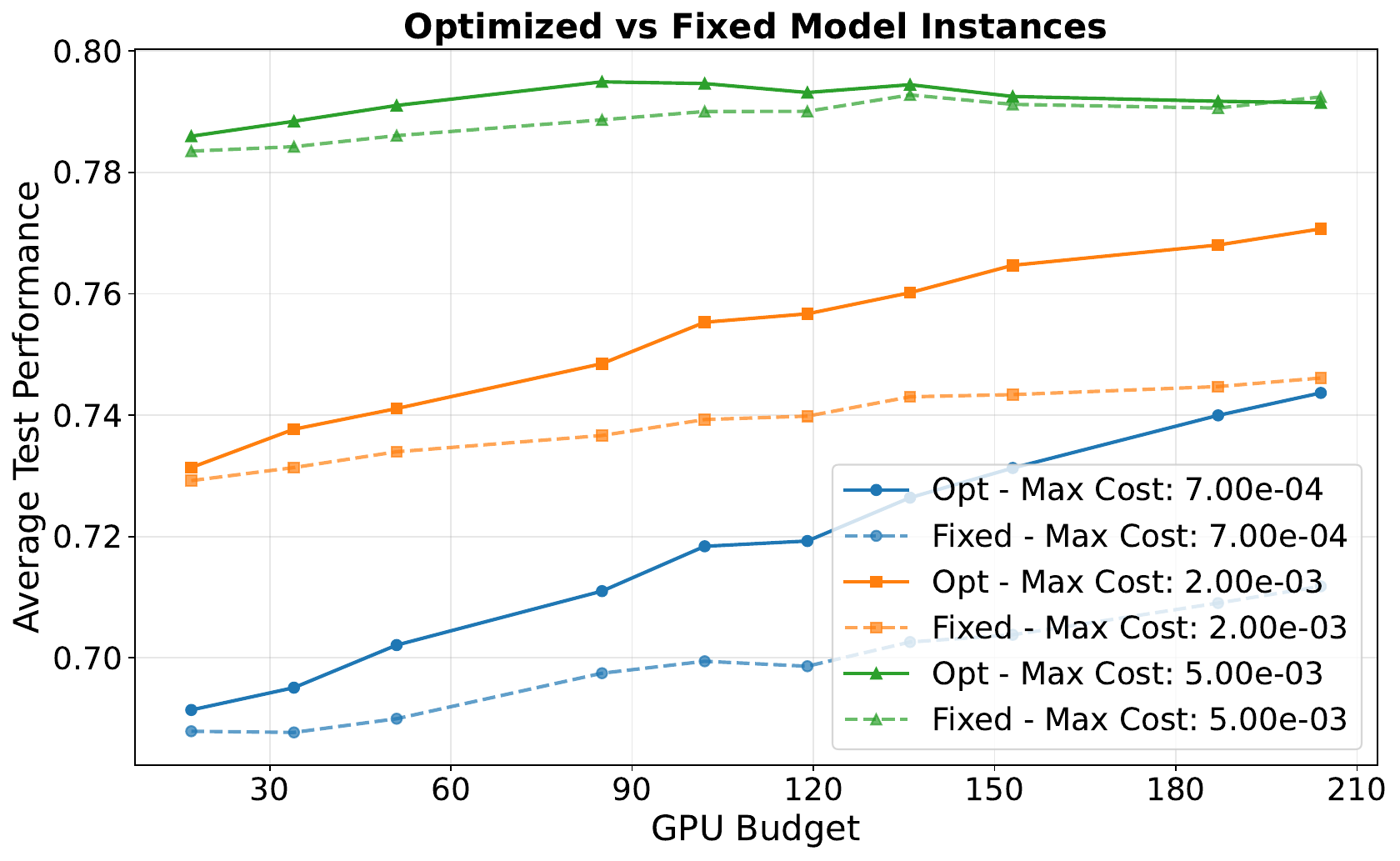}
    \caption{Dataset 2: $B=100$}
\end{subfigure}

\caption{Average test performance versus GPU budget, comparing data-dependent (optimized) allocation with fixed (pre-specified) numbers of open-source model instances. Colors indicate different bounds on the average cost per query $C$.}
\label{fig:model:instances}
\end{figure}

Figure~\ref{fig:model:instances} illustrates the performance gains observed in simulations by optimally allocating open-source model instances compared to using a fixed, uniform number of instances across models, evaluated under three different cost budgets $C$: low, medium and high. Quantitatively, our optimization-based allocation achieves up to 2.7\% improvement in Dataset~1 and 3.2\% in Dataset~2, relative to the fixed-allocation baseline. As the cost budget for closed-source LLMs increases, the gap between fixed and optimized open-source allocations narrows. This is expected, since higher cost budgets reduce reliance on open-source models and therefore lessen the impact of how their GPU instances are allocated.

Across both datasets, we see that the optimized allocation consistently outperforms the fixed-allocation baseline, demonstrating the benefit of adapting the number of model instances to the data distribution and model characteristics. Specifically, under tighter cost budgets, the optimizer tends to prioritize smaller, more efficient models that can serve a larger portion of the batch within the GPU constraint, whereas under higher budgets it can allocate additional instances to larger, higher-capacity models, thereby increasing overall prediction quality. Moreover, the results highlight that the benefits of optimization are robust across different batch sizes and cost budgets, emphasizing the importance of a data-dependent allocation strategy when deploying multiple open-source models under limited GPU resources.

\subsection{Full Optimization with Cost and GPU Constraints}
\label{subsection:model:full}

Using the same deployment setup as in Section~\ref{subsection:model:instances}, we now solve the full optimization problem under both cost and GPU constraints. This procedure consists of an offline stage that determines the number of model instances to deploy, followed by an online batch-level optimization that maximizes batch performance. We estimate LLM performance using MIRT, XGBoost and kNN-40, as well as the robust variants of XGBoost and kNN-40 with $Q=10$. We omit kNN-5 since it is not as competitive. We fix the batch size to $B=100$ and vary the maximum allowable average batch cost per query to capture both low- and high-cost regimes for closed-source models. Additional experiments for different batch sizes $B\in \{50,200,400\}$ are provided in Appendix~\ref{subsection:full:opt:additional:exp}.

\paragraph{Performance and Efficiency of Full-Optimization Routing}

\begin{figure}[htbp]
\centering

\begin{subfigure}{\columnwidth}
    \centering
    \includegraphics[width=\columnwidth]{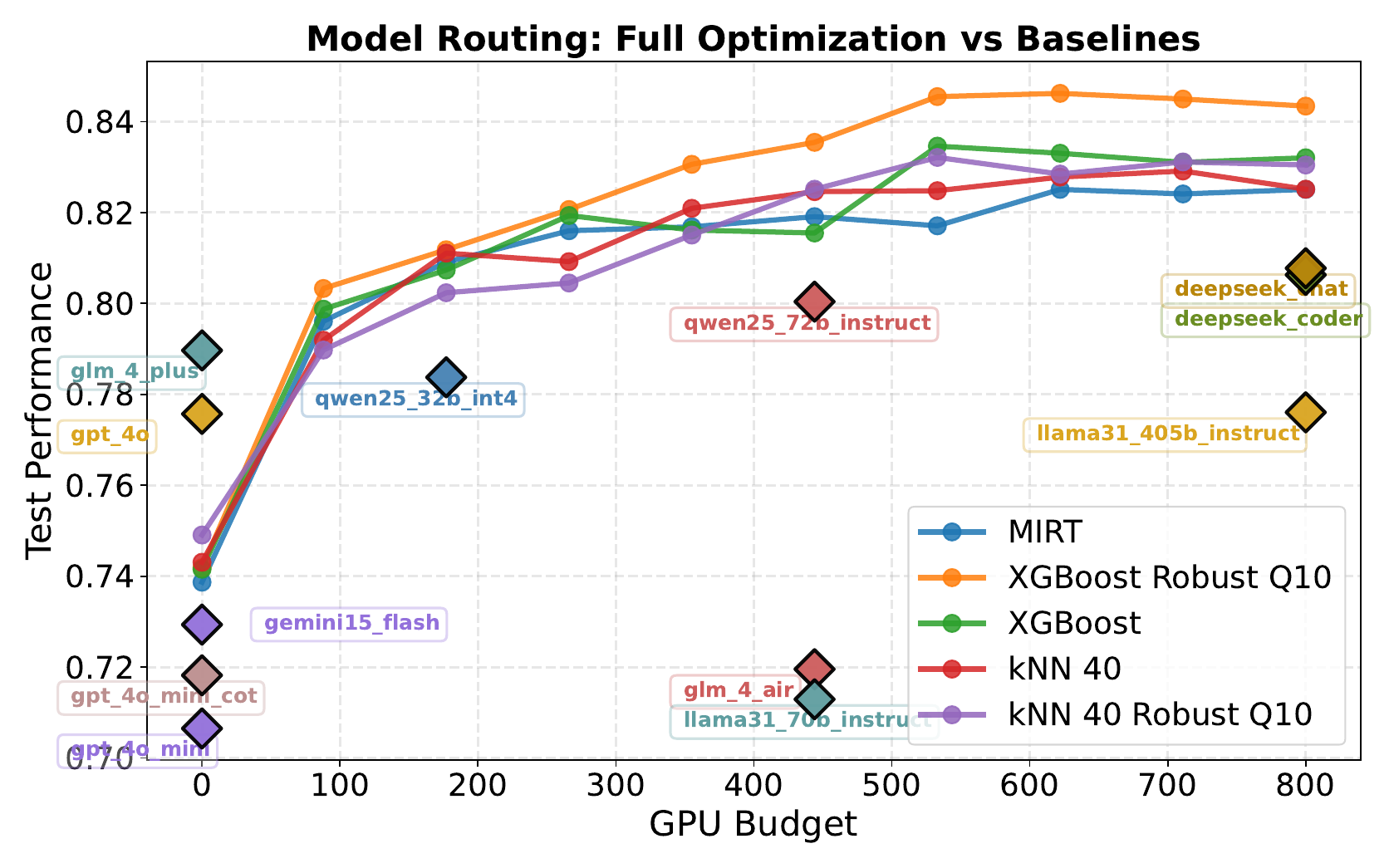}
    \caption{Dataset 1: $B=100$, $C=10^{-6}$}
\end{subfigure}

\vspace{0.5em} 

\begin{subfigure}{\columnwidth}
    \centering
    \includegraphics[width=\columnwidth]{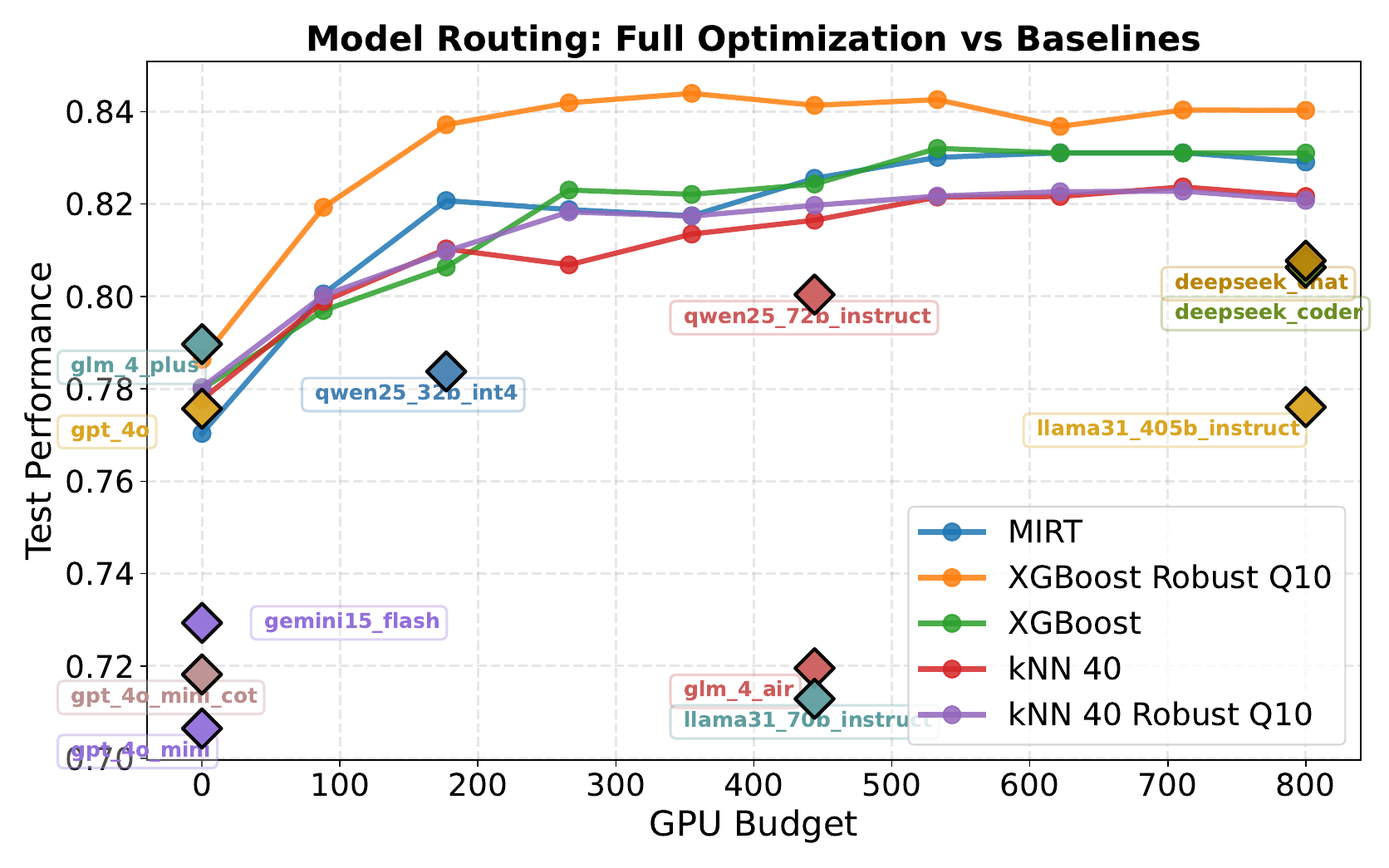}
    \caption{Dataset 1: $B=100$, $C=3\cdot 10^{-6}$}
\end{subfigure}

\caption{Average test performance as a function of the GPU budget for different LLM performance estimation methods after full optimization under cost and resource constraints on Dataset~1. We report results for both low-cost (top) and high-cost (bottom) scenarios. We also include individual LLMs with test performance above 70\%; these points correspond to benchmark-reported performance and cost values used as baselines in the simulation. Open-source models require non-zero GPU resources, while closed-source models incur no GPU requirements as per our simulation setup.}
\label{fig:full:opt:dataset:1}
\end{figure}

\begin{figure}[htbp]
\centering
\begin{subfigure}{\columnwidth}
    \centering
    \includegraphics[width=\columnwidth]{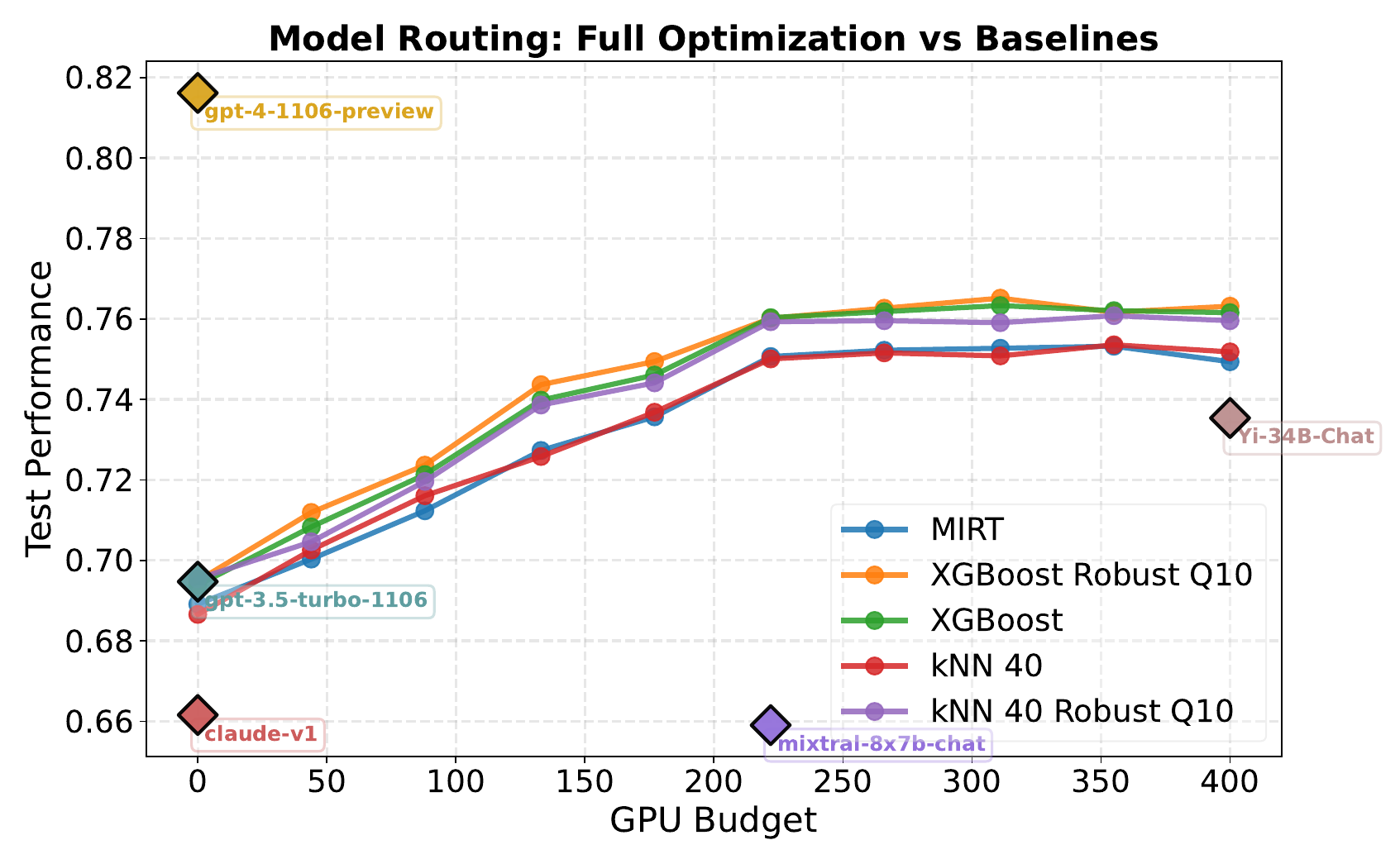}
    \caption{Dataset 2: $B=100$, $C=0.001$}
\end{subfigure}

\begin{subfigure}{\columnwidth}
    \centering
    \includegraphics[width=\columnwidth]{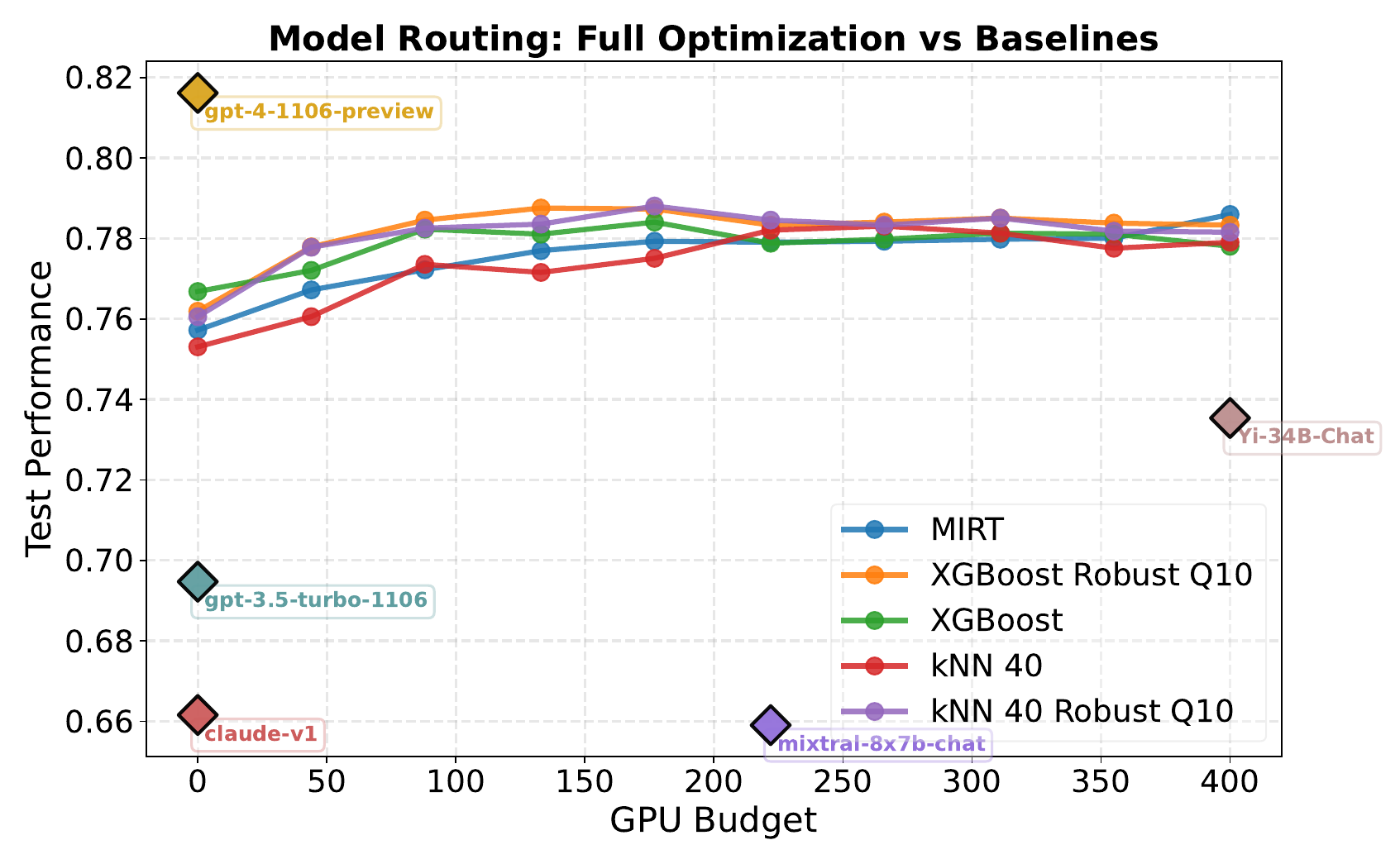}
    \caption{Dataset 2: $B=100$, $C=0.004$}
    \label{fig:full:opt:dataset:2:high:cost}
\end{subfigure}

\caption{Full optimization results on Dataset 2; see the caption of Figure~\ref{fig:full:opt:dataset:1} for a detailed description.}
\label{fig:full:opt:dataset:2}
\end{figure}

\begin{enumerate}
    \item Figure~\ref{fig:full:opt:dataset:1} presents the full optimization results on Dataset 1. The robust XGBoost $Q=10$ method achieves the best performance among all approaches, improving by up to 2\% over MIRT. Moreover, full-optimization routing reduces both GPU requirements and overall cost relative to deploying individual LLMs, while also outperforming even the strongest single model. For example, to match the performance of the top-performing model, DeepSeek\_Chat, full-optimization routing with robust XGBoost $Q=10$ requires only 177 GPUs and low cost budget of $C=10^{-6}$, compared to 800 GPUs when using DeepSeek\_Chat alone.
    \item Figure~\ref{fig:full:opt:dataset:2} reports the full-optimization results on Dataset~2 with cost and GPU constraints. The robust XGBoost $Q=10$ achieves the best performance here as well, improving by up to 1.6\% compared to MIRT. 
    We observe that robust kNN consistently outperforms its non-robust counterpart, yielding improvements of up to 1.7\%. Robust XGBoost provides smaller gains over standard XGBoost -- up to 0.2\% for $B=100$, though the improvement increases to as much as 2\% for smaller batch sizes (see Figure~\ref{fig:full:optimization:dataset:2:B:50} in the Appendix for details). Under the high-cost budget setting (Figure~\ref{fig:full:opt:dataset:2:high:cost}), the router primarily selects closed-source models; consequently, increasing the GPU budget does not lead to further performance gains.

\end{enumerate}

The conclusions regarding robustness are consistent with the per-query routing results in Section~\ref{subsection:robust:opt:per:query}: incorporating robustness into an existing performance estimator improves performance in some cases and does not degrade it in others.
These results show that full optimization -- integrating robust performance estimation with cost- and GPU-aware decision-making -- consistently improves both accuracy and resource efficiency. Beyond boosting predictive performance, the approach offers a systematic way to convert model-level performance estimates into effective system-level deployment and routing decisions, making it well suited for large-scale LLM deployments in practice.

\section{Limitations}

While our framework provides a clear formulation with strong empirical results, it departs from real-world deployment settings.

\begin{itemize}

\item \textbf{Batching assumptions.} We optimize over fixed batches, whereas production systems operate under continuous, dynamic request streams \citep{kwon2023efficient}. Batches may be only partially observed, and routing decisions must often be made online.

\item \textbf{Distribution shift.} Our evaluation assumes a stationary distribution. In practice, query distributions evolve, which can degrade both performance estimates $a_{i,j}$ and their uncertainty. Addressing this requires continual monitoring and retraining.

\item \textbf{Static evaluation.} Results are based on offline benchmarks with fixed responses, scores, and costs. Real-world performance may differ due to deployment effects and frequent model updates.

\item \textbf{Multi-turn interactions.} We assume independent queries, but real usage is conversational. Routing across models incurs context-switching overhead (no shared KV cache), which can reduce cost savings and introduce inconsistencies.

\item \textbf{Static allocation.} Model instance allocation is fixed offline and may be suboptimal under fluctuating traffic, leading to idle resources or bottlenecks. Dynamic reallocation is costly due to model loading overhead.

\end{itemize}

\section{Conclusion}

This work revisits LLM routing in practical inference systems, where queries arrive in dynamic batches under monetary and hardware constraints. Prior per-query methods exhibit high batch-level cost variability and struggle to control worst-case spending, especially under adversarial batching. To address this, we propose a batch-level routing framework based on integer linear programming that maximizes average batch performance while enforcing cost and capacity constraints. Robustness is achieved by optimizing lower bounds of the prediction intervals via bootstrap resampling, which improves performance under high uncertainty and favors models with lower predictive variance.

Batch-level optimization consistently outperforms per-query routing in both random and adversarial batches, with significant gains in worst-case scenarios. Optimizing GPU allocation across open-source models further enhances performance under fixed compute budgets. Under joint cost and GPU constraints, our approach surpasses state-of-the-art baselines such as MIRT. These results underscore that LLM routing benefits from a constrained batch-level optimization framework, enabling stable, cost- and resource-aware deployment while supporting extensions such as latency targets, token-based pricing, and adaptive uncertainty under distribution shifts.

\section*{Acknowledgement}

Rahul Mazumder contributed to this work while he was a consultant for LinkedIn (in compliance with MIT's outside professional activities policies).

\bibliographystyle{ACM-Reference-Format}
\bibliography{ref}

\appendix

\onecolumn
\section{Additional Experiment Details} \label{appendix:experimental:details}

\subsection{LLM Costs and Profiles} \label{subsection:llm:pricing:profiles}

We report the LLM costs and profiles used in the per-query optimization setting without resource constraints:
\begin{enumerate}
\item For Dataset 1, we use Table~\ref{table:llm:pricing:dataset:1}, reproduced from \citet{song2025irt}, which reports the price per million input and output tokens for each LLM. In our per-query optimization, we use the output cost per 1M tokens as the LLM cost. 
The corresponding LLM profiles are taken directly from \citet{song2025irt} and are therefore omitted here.
\item For Dataset 2, we use Table~\ref{table:llm:pricing:dataset:2}, which reports the average per-query cost for each LLM. The individual costs for each query–LLM pair are taken from~\citep{nearestneigbor}. These values are averaged and used as the LLM costs in the per-query optimization. The LLM profiles for Dataset 2 are generated using ChatGPT 5.2, following the same structure as those provided for Dataset 1, and are reported in Table~\ref{table:llm:profiles:dataset:2}.
\end{enumerate}

It is worth noting that the costs used here correspond to a snapshot in time and may differ from current pricing.

\begin{table}[ht]
\centering
\begin{tabular}{lcc}
\toprule
\textbf{LLM} & \textbf{Input \$/1M tokens} & \textbf{Output \$/1M tokens} \\
\midrule
DeepSeek-Chat & 0.14 & 0.28 \\
DeepSeek-Coder & 0.14 & 0.28 \\
Gemini-1.5-Flash & 0.075 & 0.30 \\
GLM-4-Air & 0.137 & 0.137 \\
GLM-4-Flash & 0.0137 & 0.0137 \\
GLM-4-Plus & 6.85 & 6.85 \\
GPT-4o & 2.5 & 10 \\
GPT-4o Mini & 0.15 & 0.6 \\
GPT-4o Mini + CoT & 0.15 & 0.6 \\
Llama3.1-8B-Instruct & 0.1 & 0.2 \\
Llama3.1-70B-Instruct & 0.792 & 0.792 \\
Llama3.1-405B-Instruct & 3.15 & 3.15 \\
Ministral-8B-Instruct-2410 & 0.1 & 0.2 \\
Mistral-7B-Instruct-v0.2 & 0.1 & 0.2 \\
Mixtral-8x7B-Instruct & 0.54 & 0.54 \\
Qwen2.5-32B-Instruct-GPTQ-Int4 & 0.1 & 0.2 \\
Qwen2.5-7B-Instruct & 0.1 & 0.2 \\
Qwen2.5-72B-Instruct & 1.08 & 1.08 \\
Qwen2.5-Math-7B-Instruct & 0.1 & 0.2 \\
QwQ-32B-Preview & 1.2 & 1.2 \\
\bottomrule
\end{tabular}
\caption{Dataset 1: Pricing of LLMs in USD per million tokens taken directly from \citet{song2025irt}.}
\label{table:llm:pricing:dataset:1}
\end{table}

\begin{table}[htbp]
\centering
\begin{tabular}{l r}
\hline
\textbf{LLM} & \textbf{Average Cost Per Query} \\
\hline
WizardLM-13B-V1.2 & 0.0001417 \\
Claude-Instant-V1 & 0.0012362 \\
Claude-V1 & 0.0058702 \\
Claude-V2 & 0.0061528 \\
GPT-3.5-Turbo-1106 & 0.0007086 \\
GPT-4-1106-Preview & 0.0079425 \\
Code-Llama-Instruct-34B-Chat & 0.0005499 \\
Llama-2-70B-Chat & 0.0013373 \\
Mistral-7B-Chat & 0.0001388 \\
Mixtral-8x7B-Chat & 0.0004140 \\
Yi-34B-Chat & 0.0005579 \\
\hline
\end{tabular}
\caption{Dataset 2: Average cost per query for each model, with the individual costs for each query-LLM pair taken directly from \citet{nearestneigbor}.}
\label{table:llm:pricing:dataset:2}
\end{table}

\begin{longtable}{p{0.28\textwidth} p{0.68\textwidth}}

\hline
\textbf{LLM} & \textbf{Profile / Description} \\
\hline
\endfirsthead

\hline
\textbf{Model} & \textbf{Profile / Description} \\
\hline
\endhead

\hline
\endfoot

\endlastfoot

gpt-3.5-turbo-1106 &
Released by OpenAI in November 2023, GPT-3.5-turbo-1106 is a proprietary chat-optimized large language model in the GPT-3.5 family, designed for efficient conversational and general text generation tasks. It supports function calling and improved JSON-structured outputs, and is trained on a mixture of licensed data, human-created data, and publicly available text. \\

claude-instant-v1 &
Released by Anthropic in 2023, Claude-Instant-v1 is a proprietary lightweight conversational language model designed for low-latency and cost-efficient deployments. It emphasizes fast response times, summarization, document analysis, and general dialogue capabilities, while maintaining Anthropic’s ``helpful, honest, and harmless'' alignment philosophy. \\

claude-v1 &
Introduced by Anthropic in March 2023, Claude-v1 is a proprietary large language model designed for conversational AI, reasoning, summarization, and instruction following. It was among the first models trained using Anthropic’s constitutional AI framework, prioritizing safety and controllability. \\

claude-v2 &
Released by Anthropic in mid-2023, Claude-v2 improves upon Claude-v1 with stronger reasoning, coding ability, and longer context handling. It supports larger context windows and delivers improved performance on math, logic, and complex instruction-following tasks. \\

gpt-4-1106-preview &
Released by OpenAI in November 2023, GPT-4-1106-preview is a proprietary preview model from the GPT-4 Turbo family, optimized for faster inference and lower cost while retaining GPT-4-level reasoning. It supports function calling and improved JSON output reliability. \\

llama-2-70b-chat &
Released by Meta AI in 2023, LLaMA-2-70B-Chat is an open-weights 70B-parameter model fine-tuned for dialogue using supervised fine-tuning and reinforcement learning from human feedback. It is widely adopted due to open licensing and competitive performance. \\

mixtral-8x7b-chat &
Released by Mistral AI in early 2024, Mixtral-8$\times$7B-Chat is an open-weights sparse Mixture-of-Experts model that activates a subset of experts per token, enabling strong performance with improved inference efficiency. \\

Yi-34B-Chat &
Released by 01.AI in early 2024, Yi-34B-Chat is an open-weights 34B-parameter bilingual (English/Chinese) language model optimized for conversational use. It demonstrates strong reasoning, math, and general knowledge performance. \\

WizardLM-13B-V1.2 &
Released by the WizardLM team in 2023, WizardLM-13B-V1.2 is an open-weights instruction-tuned model trained using an ``evolutionary instruction'' approach that generates diverse instruction data, yielding strong instruction-following performance for its size. \\

code-llama-instruct-34b-chat &
Released by Meta AI in 2024, Code Llama Instruct 34B Chat is an open-weights model specialized for programming and code-centric dialogue. It is fine-tuned for code generation, debugging, and explanation across multiple programming languages. \\

mistral-7b-chat &
Released by Mistral AI in late 2023, Mistral-7B-Chat is an open-weights chat-tuned model with architectural improvements such as grouped-query attention and sliding-window attention, enabling efficient inference and long-context handling. \\
\hline
\caption{Dataset 2: Profiles of the language models used in experiments.}
\label{table:llm:profiles:dataset:2}
\end{longtable}

\subsection{Performance of Robust Methods Across Quantiles} \label{subsection:per:query:quantiles}

\begin{figure}[htbp]
\centering
\begin{subfigure}[b]{0.8\textwidth}
    \includegraphics[width=\textwidth]{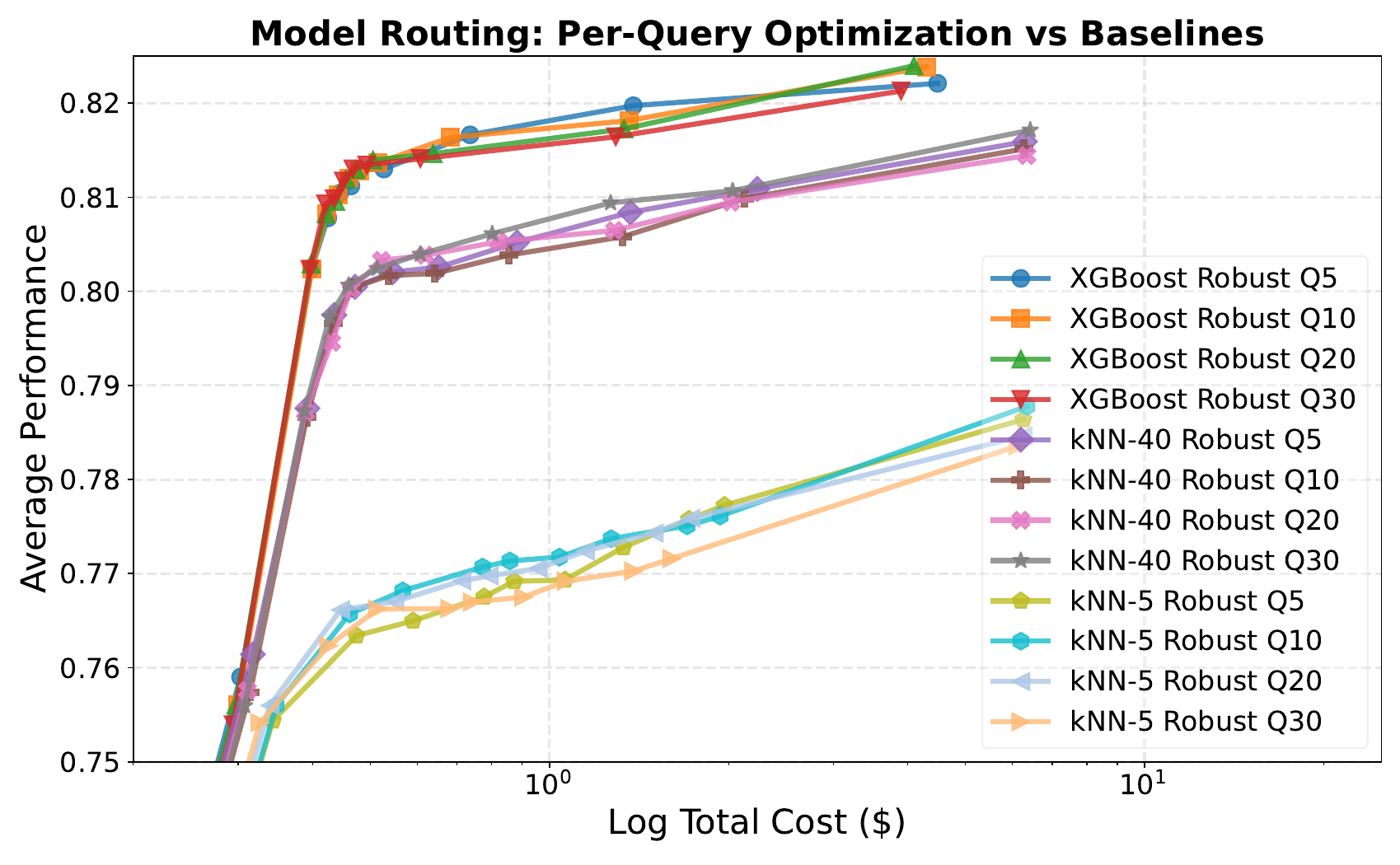}
    \caption{Dataset 1}
\end{subfigure}

\begin{subfigure}[b]{0.8\textwidth}
    \includegraphics[width=\textwidth]{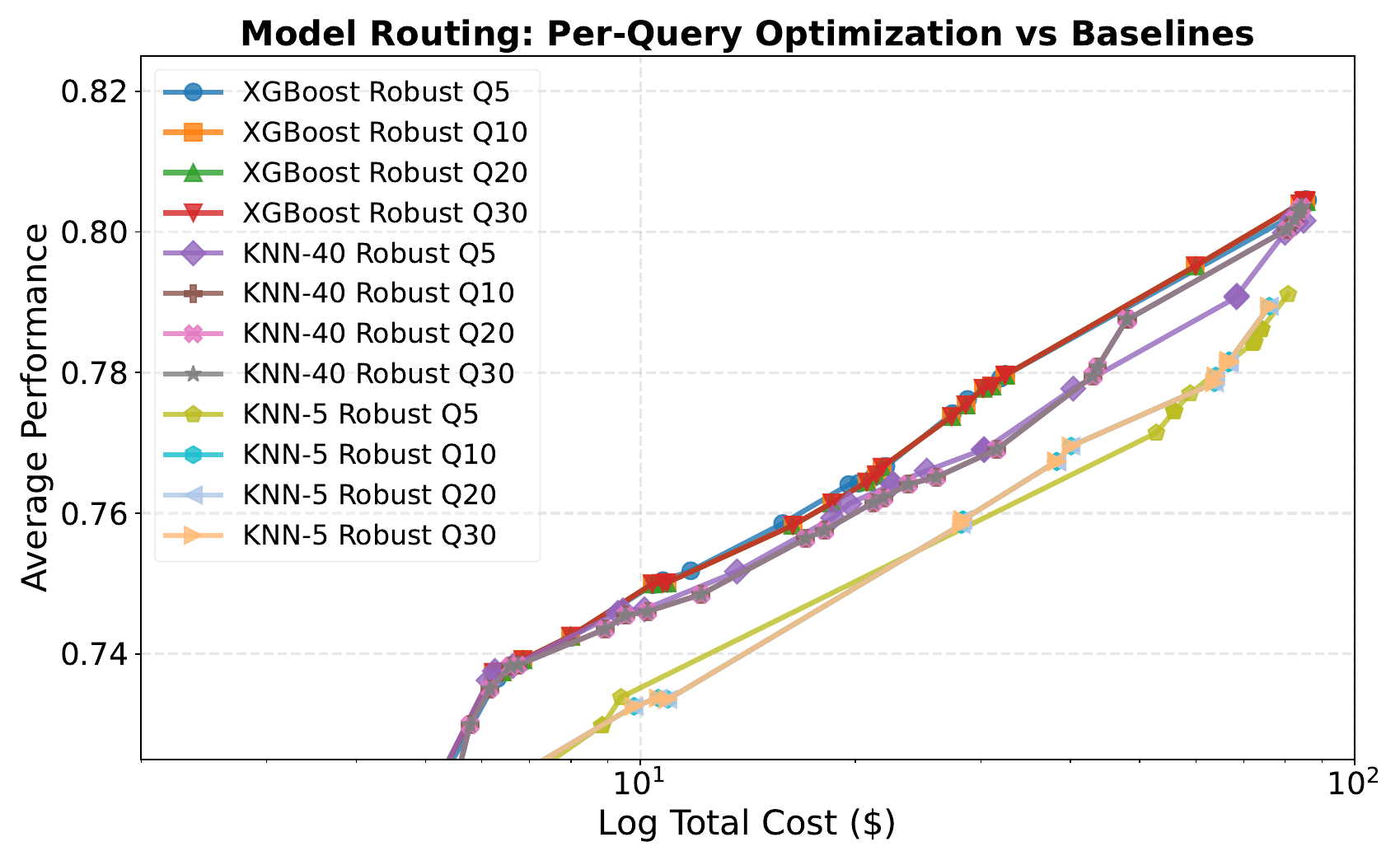}
    \caption{Dataset 2}
\end{subfigure}

\caption{Performance versus cost of robust versions of kNN-5, kNN-40 and XGBoost under per-query optimization across different quantile levels $Q\in \{5,10,20,30\}$.}
\label{fig:per:query:quantiles}
\end{figure}

We find that the routing performance of various robust performance estimators exhibits only a weak sensitivity to the choice of quantile. Using the experimental setup described in Section~\ref{subsection:robust:opt:per:query}, Figure~\ref{fig:per:query:quantiles} illustrates that, across both datasets, the performance of each individual robust estimator kNN-5, kNN-40 and and XGBoost remains largely consistent when the quantile $Q$ is varied over the range $\{5,10,20,30\}$. This suggests that the effectiveness of these robust methods is not strongly tied to a particular quantile selection, highlighting their stability and practical flexibility in routing scenarios.

\subsection{Batch-Level Optimization for Worst-Case Cost Control: Additional Experiments} \label{subsection:cost:additional:exp}

\begin{figure}[htbp]
\centering
\begin{subfigure}[b]{1.0\textwidth}
    \includegraphics[width=\textwidth]{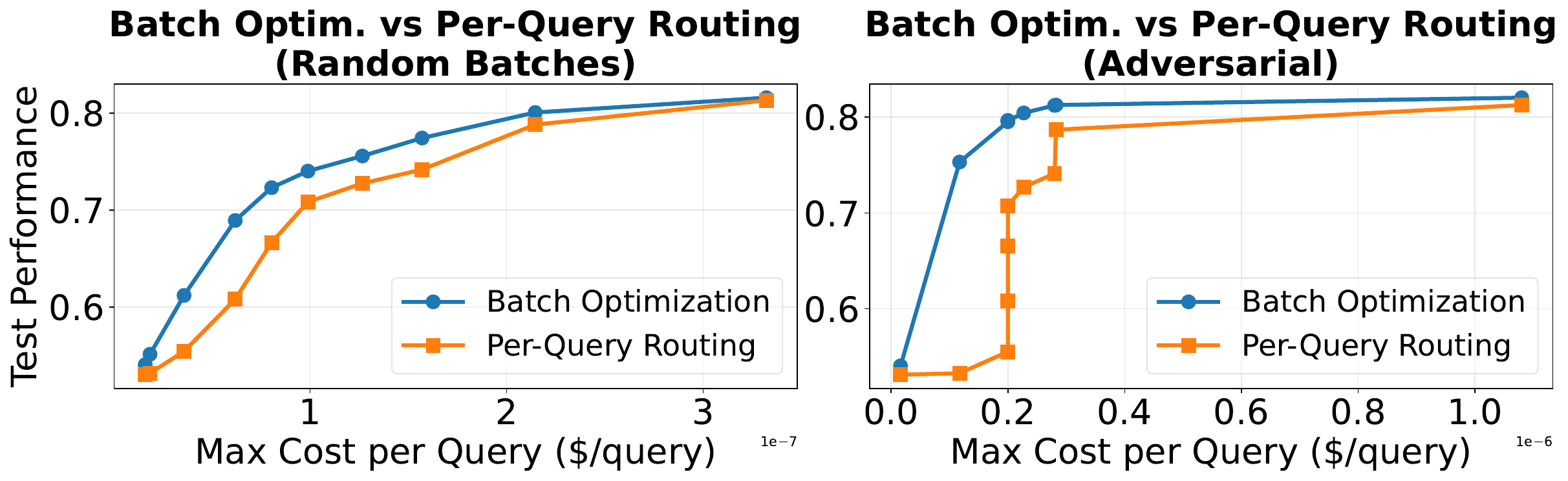}
    \caption{Dataset 1: $B=50$}
\end{subfigure}

\begin{subfigure}[b]{1.0\textwidth}
    \includegraphics[width=\textwidth]{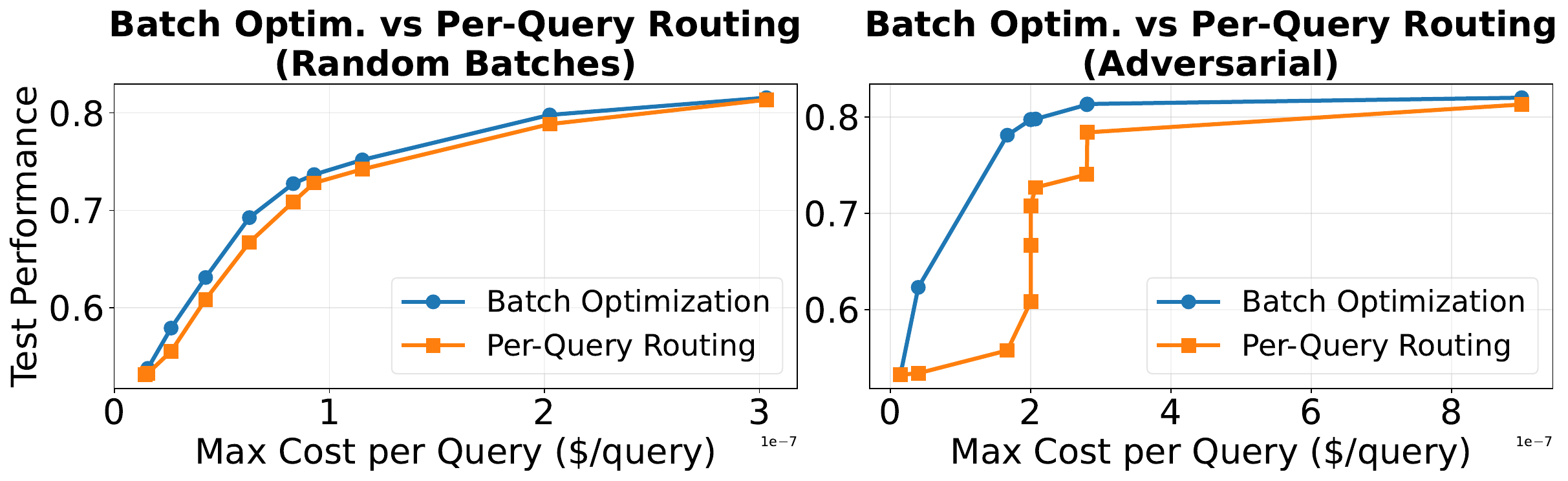}
    \caption{Dataset 1: $B=200$}
\end{subfigure}

\begin{subfigure}[b]{1.0\textwidth}
    \includegraphics[width=\textwidth]{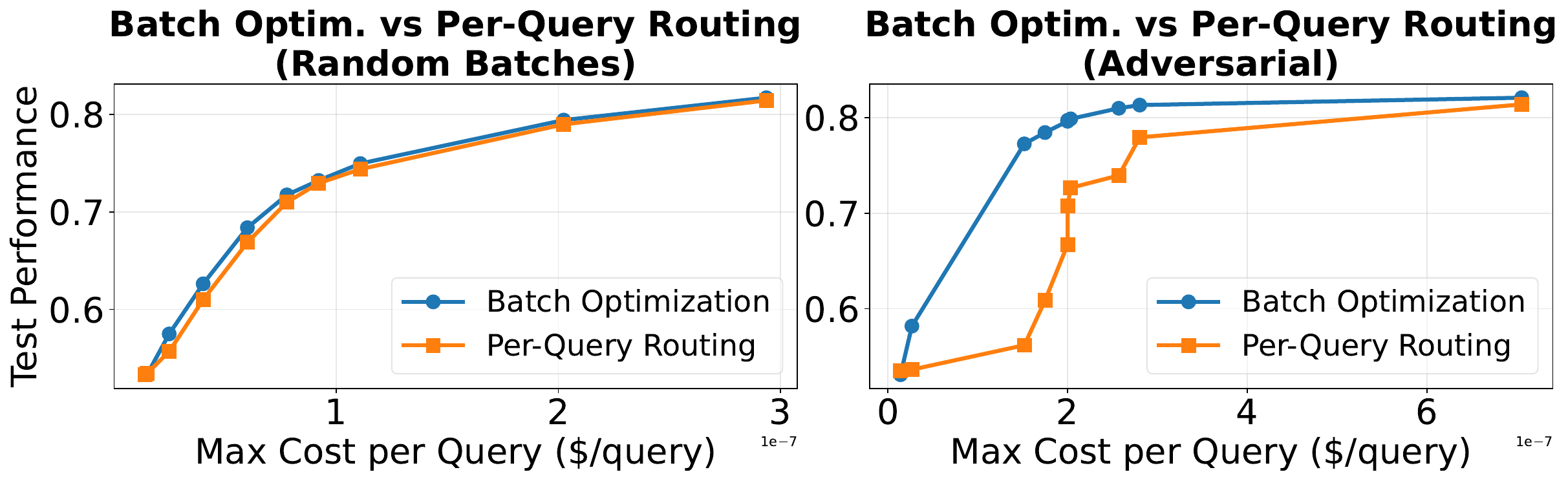}
    \caption{Dataset 1: $B=400$}
\end{subfigure}
\caption{Average test performance vs.~maximum batch-level average per-query cost across batches, comparing per-query optimization with batch-level optimization. Results are reported for Dataset 1 using robust XGBoost Q10 as the performance estimator, under randomly constructed batches (left) and adversarial batches (right).}
\label{fig:cost:additional:dataset:1}
\end{figure}

\begin{figure}[htbp]
\centering
\begin{subfigure}[b]{1.0\textwidth}
    \includegraphics[width=\textwidth]{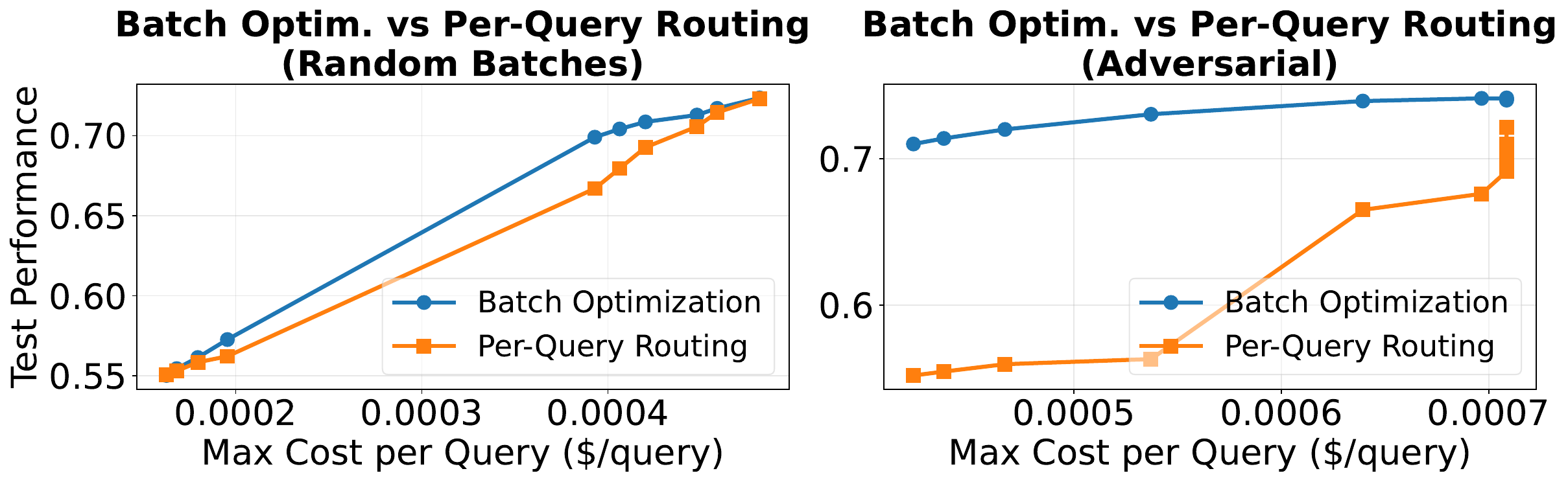}
    \caption{Dataset 2: $B=50$}
\end{subfigure}

\begin{subfigure}[b]{1.0\textwidth}
    \includegraphics[width=\textwidth]{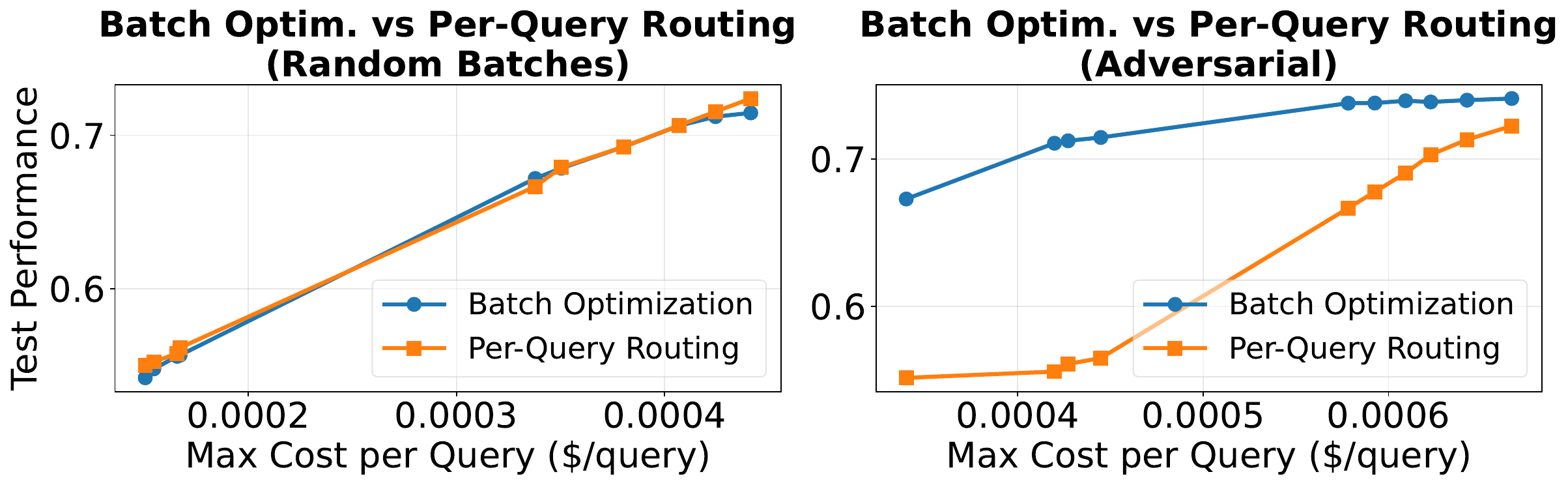}
    \caption{Dataset 2: $B=200$}
\end{subfigure}

\begin{subfigure}[b]{1.0\textwidth}
    \includegraphics[width=\textwidth]{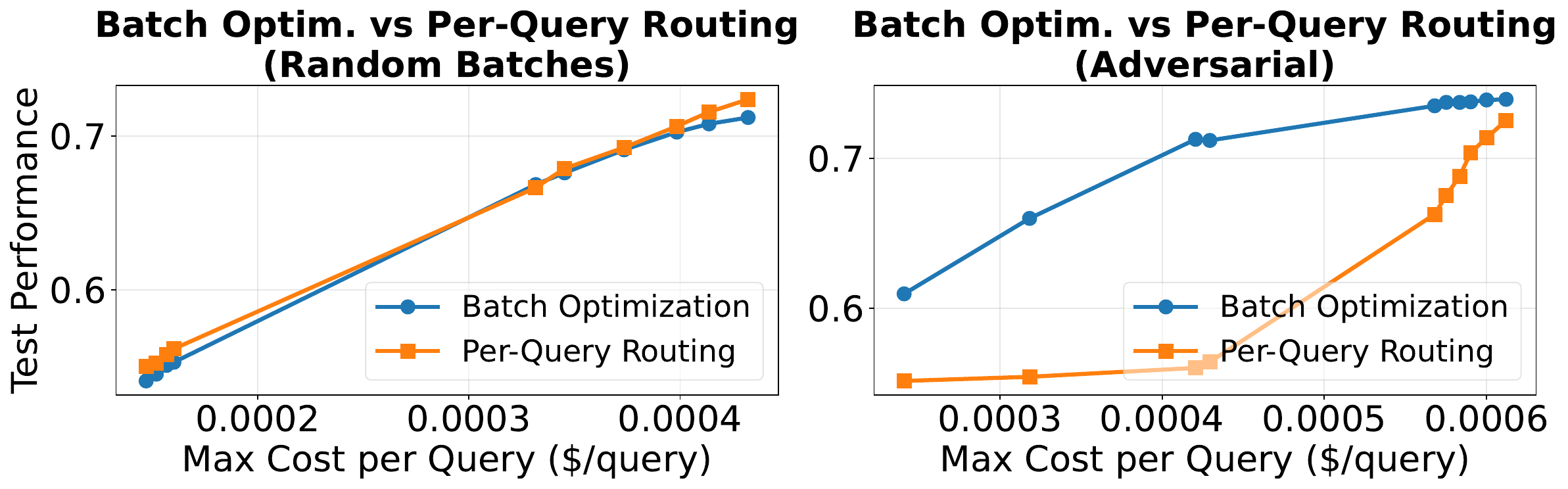}
    \caption{Dataset 2: $B=400$}
\end{subfigure}
\caption{Average test performance vs.~maximum batch-level average per-query cost across batches, comparing per-query optimization with batch-level optimization. Results are reported for Dataset 2 using robust XGBoost $Q=10$ as the performance estimator, under randomly constructed batches (left) and adversarial batches (right).}
\label{fig:cost:additional:dataset:2}
\end{figure}

Using the same experimental setup from Section~\ref{subsection:model:batch-level}, Figures~\ref{fig:cost:additional:dataset:1} and~\ref{fig:cost:additional:dataset:2} provide additional comparisons of batch-level versus per-query optimization for batch sizes $B \in \{50, 200, 400\}$ on Dataset 1 and Dataset 2, respectively. 

\subsection{LLM Experimental Deployment Setup}\label{subsection:llm:pricing:gpus}

Table~\ref{table:llm:pricing:gpus} summarizes our experimental deployment setup, in which open-source LLMs incur zero inference cost but require GPU resources, whereas closed-source models have nonzero inference costs and no GPU requirements.

\begin{table}[htbp]
\centering

\begin{subtable}{\textwidth}
\centering
\begin{tabular}{lcc}
\toprule
\textbf{LLM} & \textbf{Output \$/1M tokens} & \textbf{\# GPU} \\
\midrule
DeepSeek-Chat & 0.0 & 8\\
DeepSeek-Coder & 0.0 & 8\\

GLM-4-Air & 0.0 & 4 \\
GLM-4-Flash & 0.0 & 1 \\

Llama3.1-405B-Instruct & 0.0 & 8\\
Llama3.1-70B-Instruct & 0.0 & 4\\
Llama3.1-8B-Instruct & 0.0 & 1 \\

Mixtral-8x7B-Instruct & 0.0 & 2\\
Ministral-8B-Instruct-2410 & 0.0 & 1\\
Mistral-7B-Instruct-v0.2 & 0.0 & 1\\

QwQ-32B-Preview & 0.0 & 4 \\ 
Qwen2.5-72B-Instruct & 0.0 & 4\\
Qwen2.5-32B-Instruct-GPTQ-Int4 & 0.0 & 1\\
Qwen2.5-7B-Instruct & 0.0 & 1\\
Qwen2.5-Math-7B-Instruct & 0.0 & 1\\
\midrule

Gemini-1.5-Flash & 0.30 & 0 \\
GLM-4-Plus & 6.85 & 0 \\
GPT-4o & 10 & 0\\
GPT-4o Mini  & 0.6 & 0 \\
GPT-4o Mini + CoT & 0.6 & 0 \\
\bottomrule
\end{tabular}
\caption{Dataset 1}
\end{subtable}

\vspace{0.6cm}

\begin{subtable}{\textwidth}
\centering
\begin{tabular}{l r r}
\toprule
\textbf{LLM} & \textbf{Average Per-Query Cost} & \textbf{\# GPU} \\
\midrule
Code-Llama-Instruct-34B-Chat & 0.0 & 4 \\
Llama-2-70B-Chat & 0.0 & 4 \\
Yi-34B-Chat & 0.0 & 4 \\
WizardLM-13B-V1.2 & 0.0 & 2 \\
Mixtral-8x7B-Chat & 0.0 & 2 \\
Mistral-7B-Chat & 0.0 & 1 \\
\midrule
GPT-3.5-Turbo-1106 & 0.0007086 & 0 \\
GPT-4-1106-Preview & 0.0079425 & 0 \\
Claude-Instant-V1 & 0.0012362 & 0 \\
Claude-V1 & 0.0058702 & 0 \\
Claude-V2 & 0.0061528 & 0 \\
\bottomrule
\end{tabular}
\caption{Dataset 2}
\label{tab:llm_cost_gpu}
\end{subtable}

\caption{Pricing and GPU requirements of LLMs across both datasets for the experimental deployment setup.}
\label{table:llm:pricing:gpus}

\end{table}

\subsection{LLM Instance Allocation: Additional Experiments} \label{subsection:llm:instances:additional:exp}

\begin{figure}[htbp]
\centering

\begin{subfigure}[b]{0.49\textwidth}
    \centering
    \includegraphics[width=\textwidth]{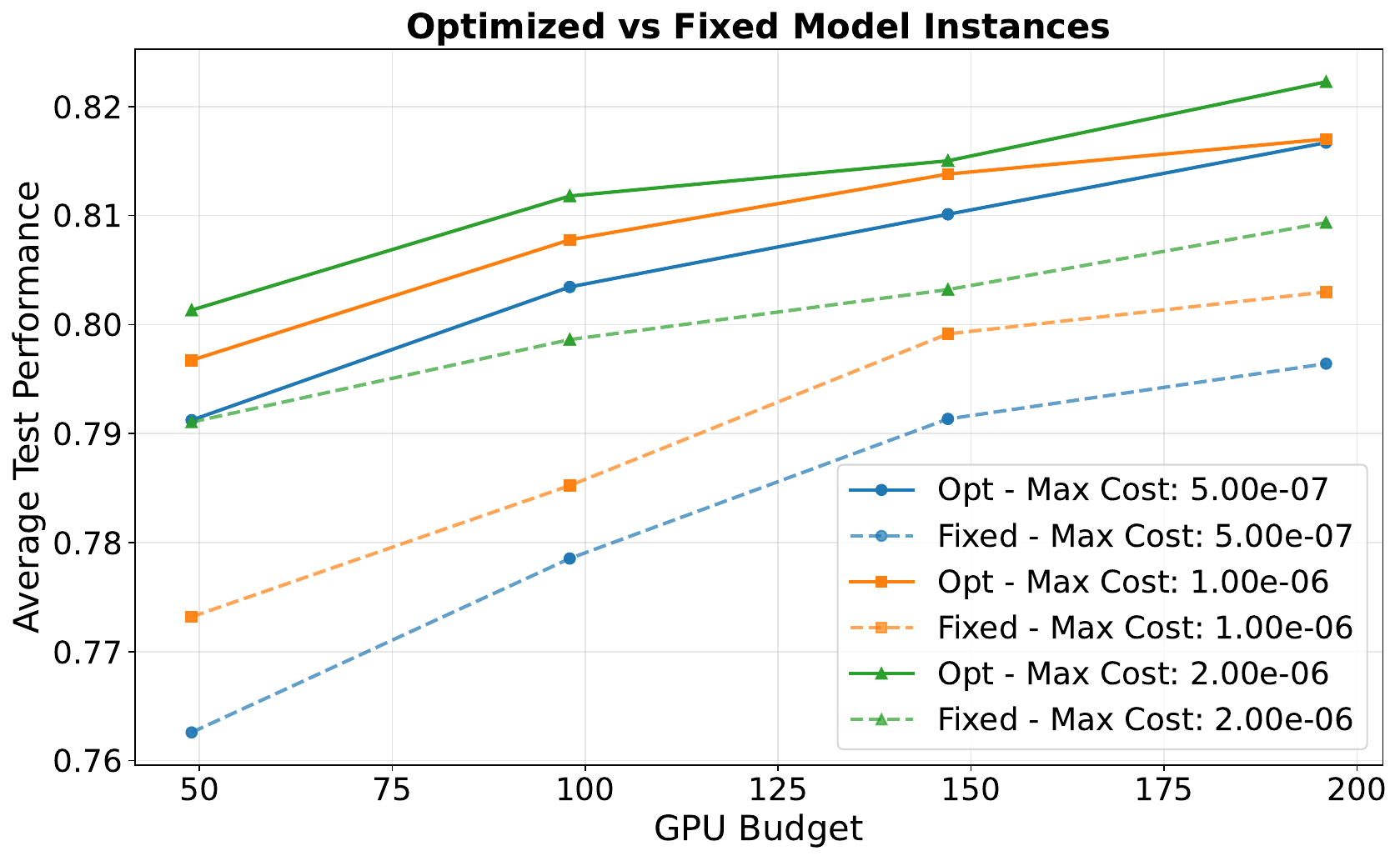}
    \caption{Dataset 1: $B=50$}
\end{subfigure}
\hfill
\begin{subfigure}[b]{0.49\textwidth}
    \centering
    \includegraphics[width=\textwidth]{figures/model_instances_test1_100.pdf}
    \caption{Dataset 1: $B=100$}
\end{subfigure}

\vspace{0.5em}

\begin{subfigure}[b]{0.49\textwidth}
    \centering
    \includegraphics[width=\textwidth]{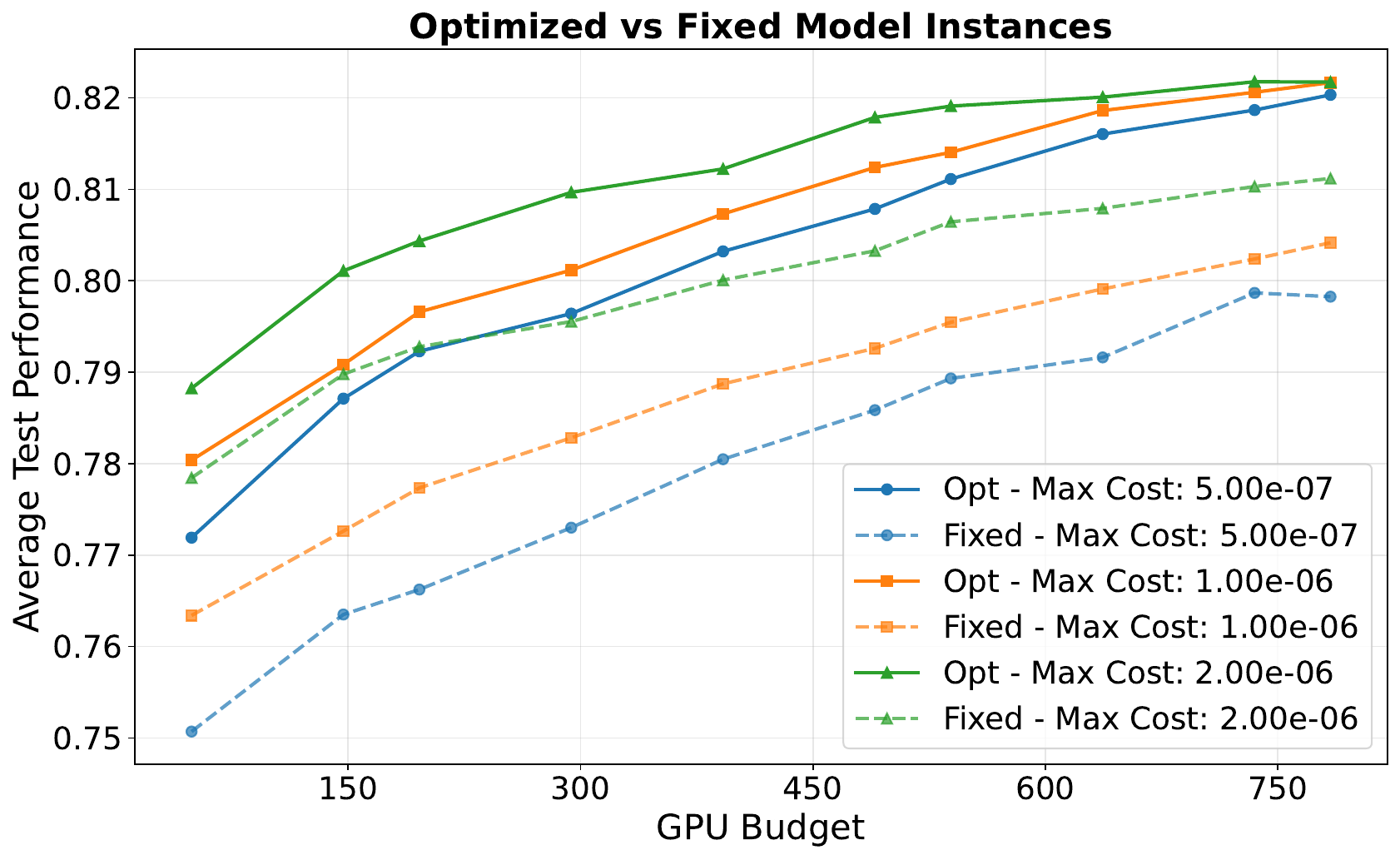}
    \caption{Dataset 1: $B=200$}
\end{subfigure}
\hfill
\begin{subfigure}[b]{0.49\textwidth}
    \centering
    \includegraphics[width=\textwidth]{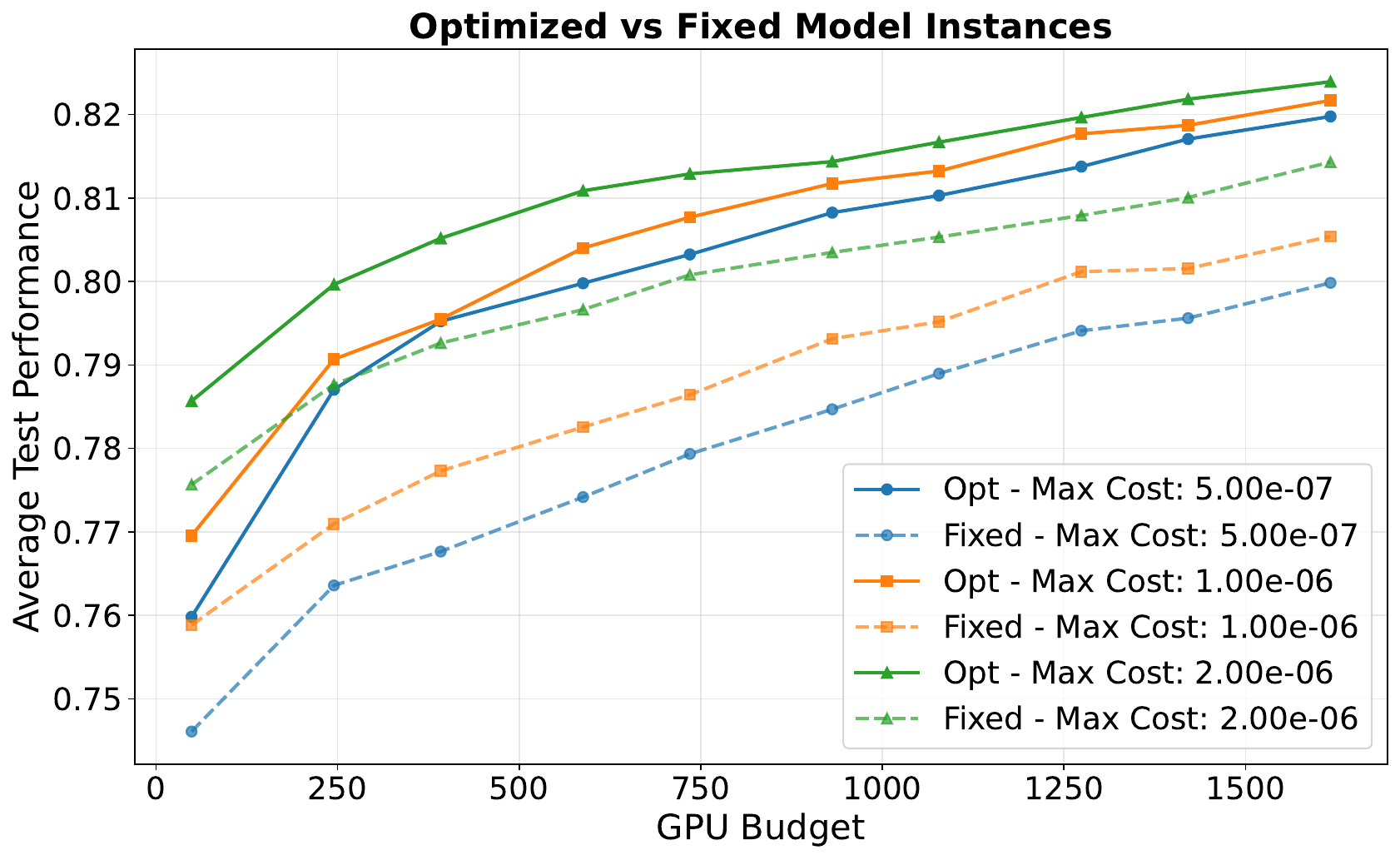}
    \caption{Dataset 1: $B=400$}
\end{subfigure}

\caption{Performance as a function of the GPU budget for optimized and fixed model instances, shown for varying batch sizes and cost budgets.}
\label{fig:model:instances:additional:dataset:1}
\end{figure}

\begin{figure}[htbp]
\centering

\begin{subfigure}[b]{0.49\textwidth}
    \centering
    \includegraphics[width=\textwidth]{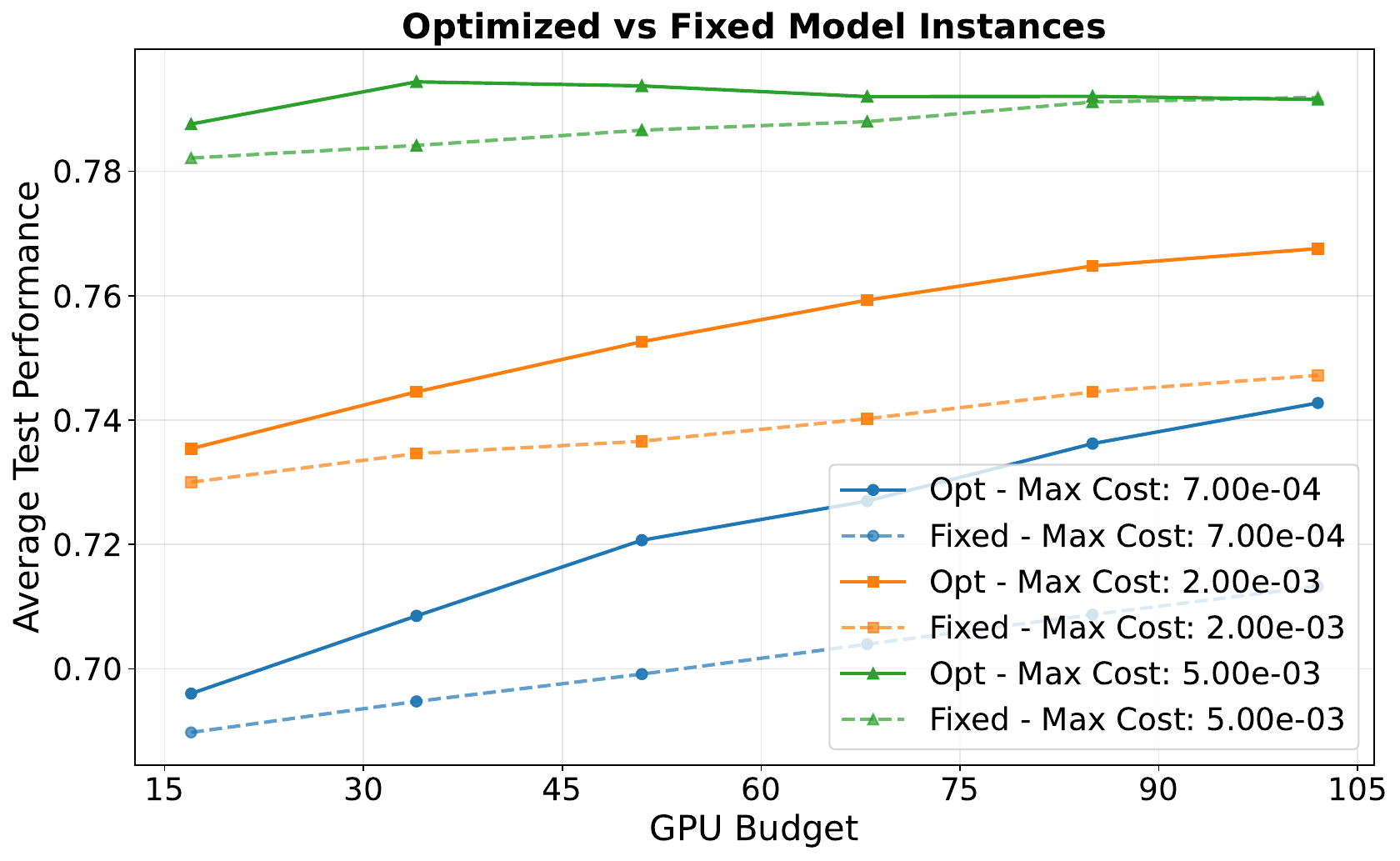}
    \caption{Dataset 2: $B=50$}
\end{subfigure}
\hfill
\begin{subfigure}[b]{0.49\textwidth}
    \centering
    \includegraphics[width=\textwidth]{figures/model_instances_rb_100.pdf}
    \caption{Dataset 2: $B=100$}
\end{subfigure}

\vspace{0.5em}

\begin{subfigure}[b]{0.49\textwidth}
    \centering
    \includegraphics[width=\textwidth]{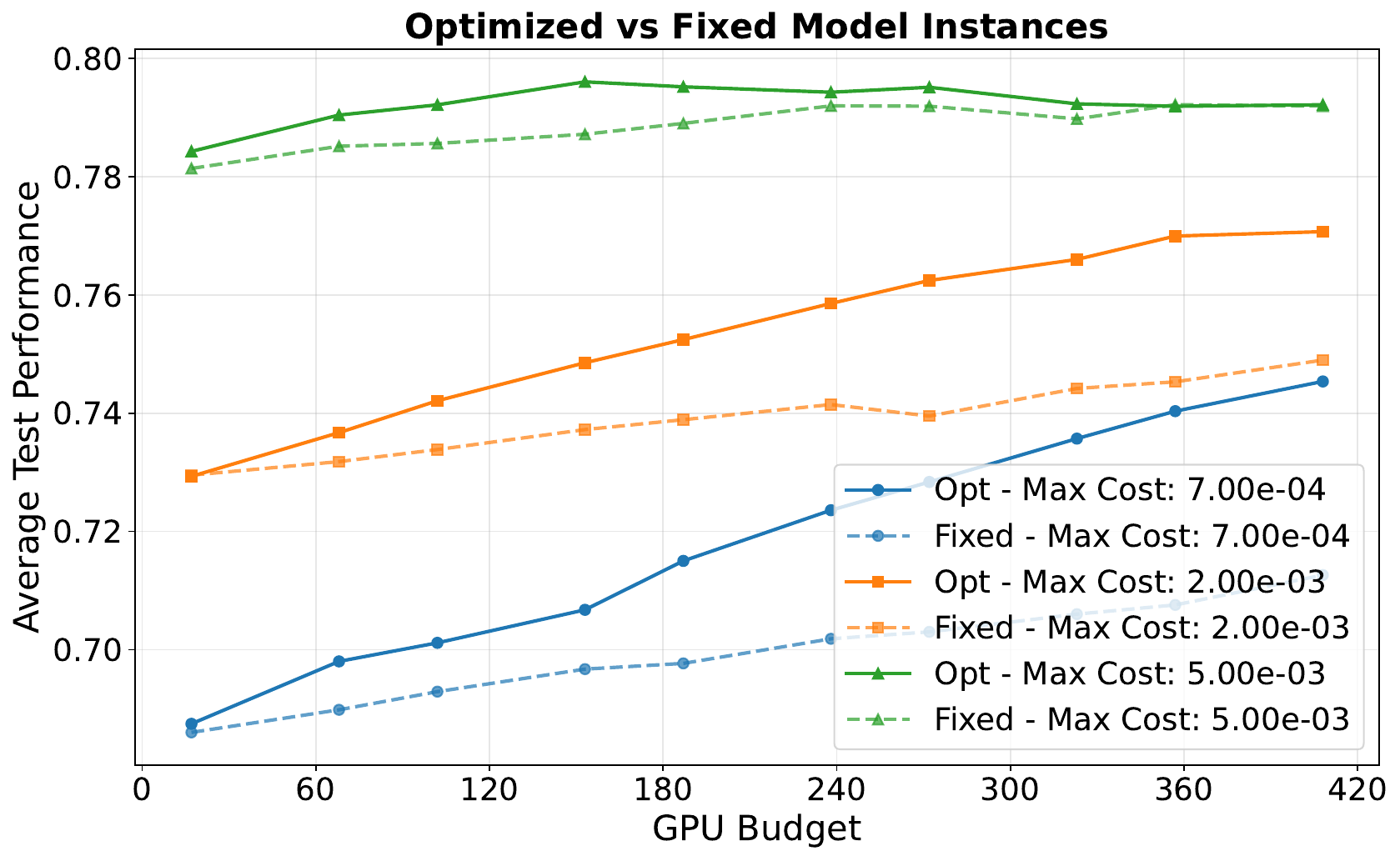}
    \caption{Dataset 2: $B=200$}
\end{subfigure}
\hfill
\begin{subfigure}[b]{0.49\textwidth}
    \centering
    \includegraphics[width=\textwidth]{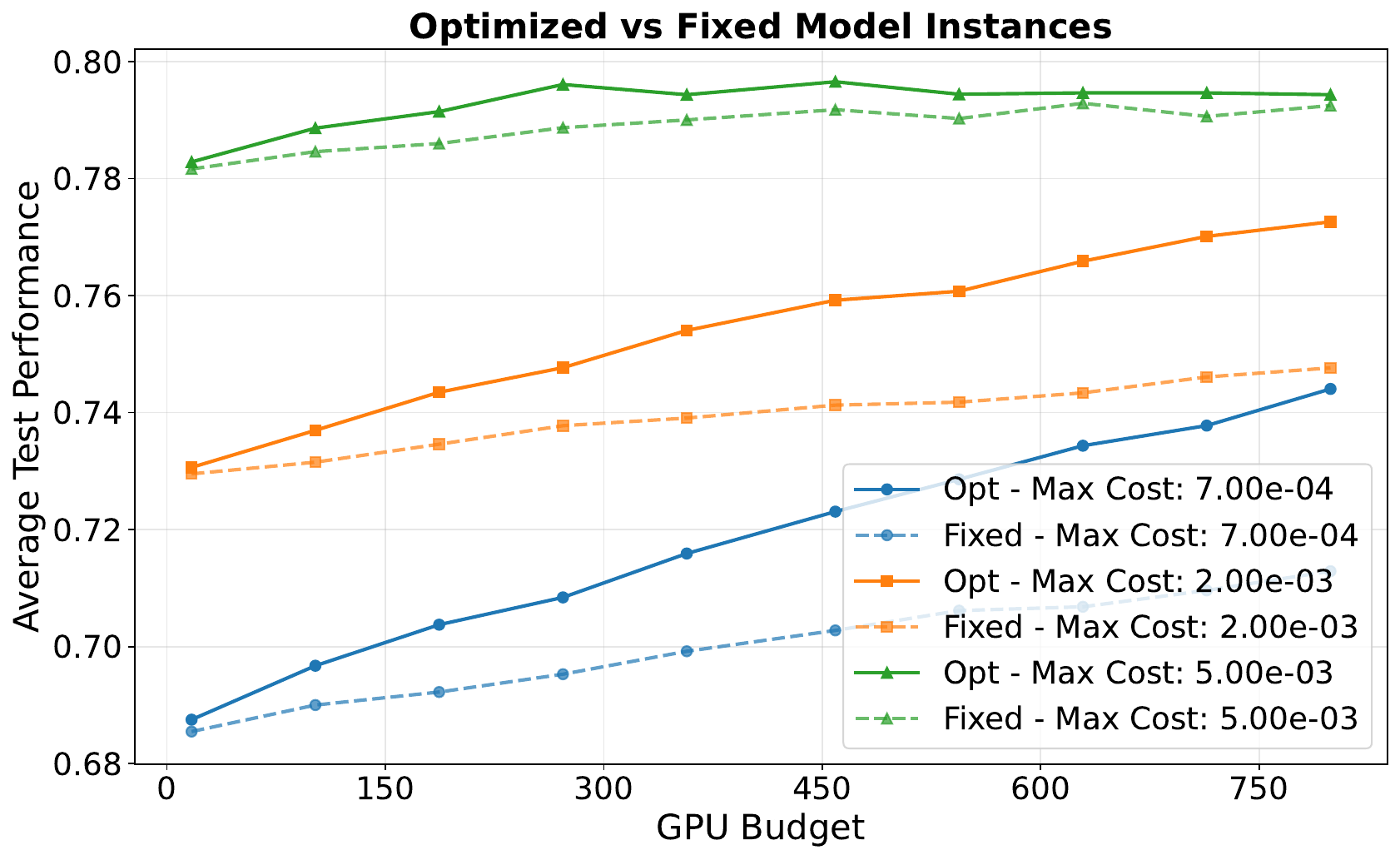}
    \caption{Dataset 2: $B=400$}
\end{subfigure}

\caption{Performance as a function of the GPU budget for optimized and fixed model instances, shown for varying batch sizes and cost budgets.}
\label{fig:model:instances:additional:dataset:2}
\end{figure}

In Figures~\ref{fig:model:instances:additional:dataset:1} and \ref{fig:model:instances:additional:dataset:2}, we report additional experiments that vary the batch size and compare performance under optimized versus fixed numbers of open-source model instances.

\subsection{Full Optimization: Additional Experiments} \label{subsection:full:opt:additional:exp}

\begin{figure}[htbp]
\centering

\begin{subfigure}[b]{0.49\textwidth}
    \centering
    \includegraphics[width=\textwidth]{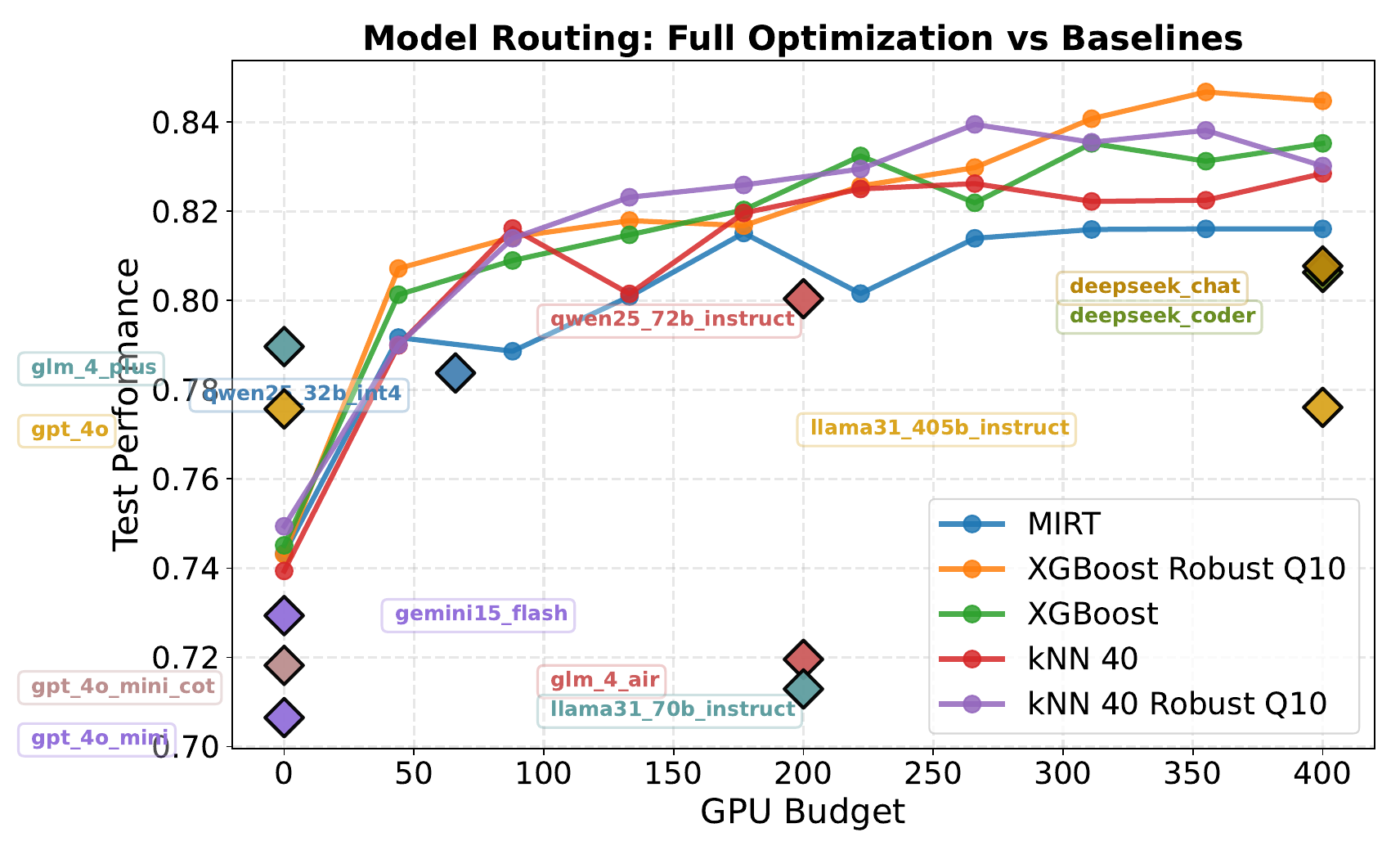}
    \caption{Dataset 1: $C=1 \cdot 10^{-6}, B=50$}
\end{subfigure}
\hfill
\begin{subfigure}[b]{0.49\textwidth}
    \centering
    \includegraphics[width=\textwidth]{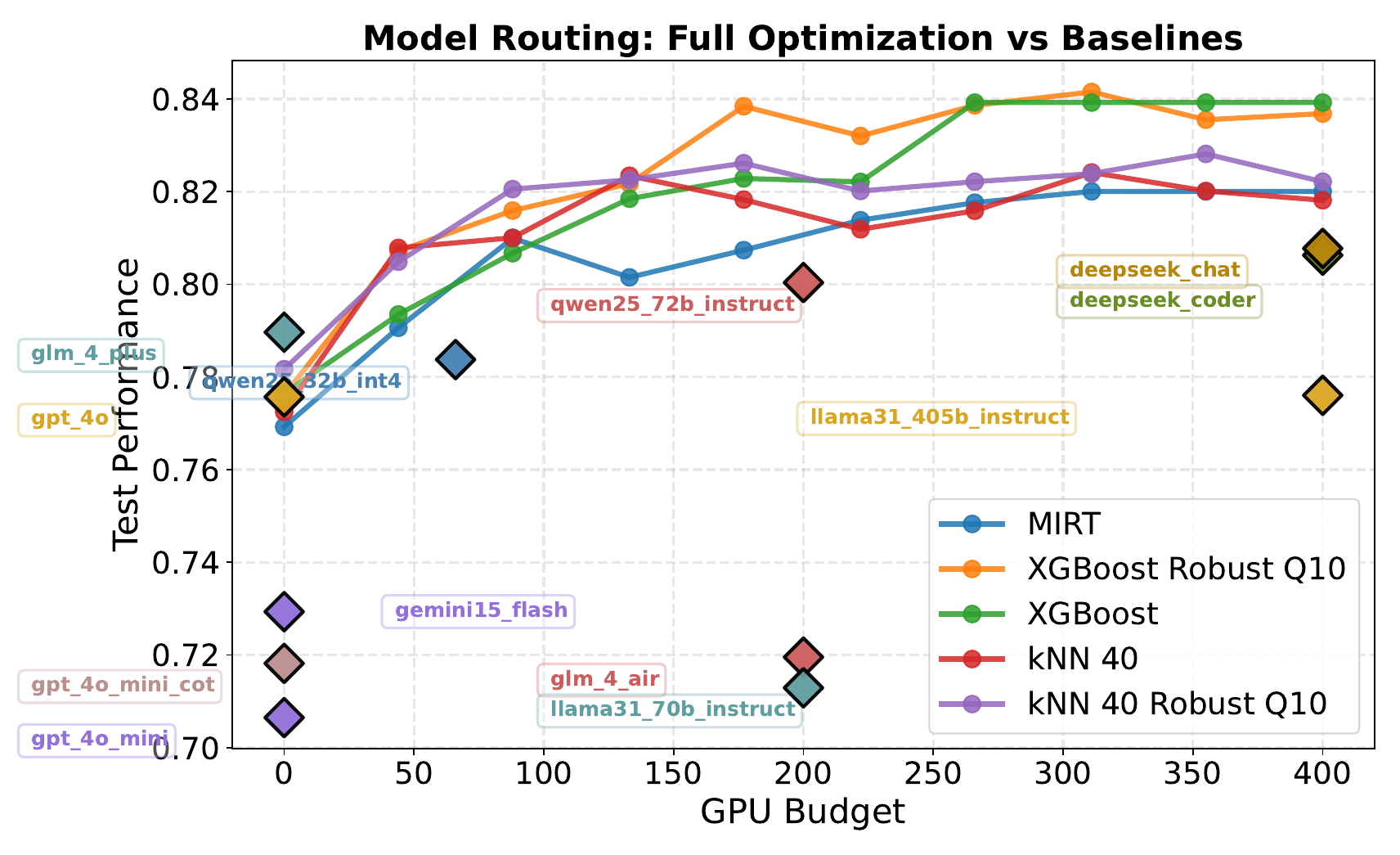}
    \caption{Dataset 1: $C=3\cdot 10^{-6}, B=50$}
\end{subfigure}

\vspace{0.5em}

\begin{subfigure}[b]{0.49\textwidth}
    \centering
    \includegraphics[width=\textwidth]{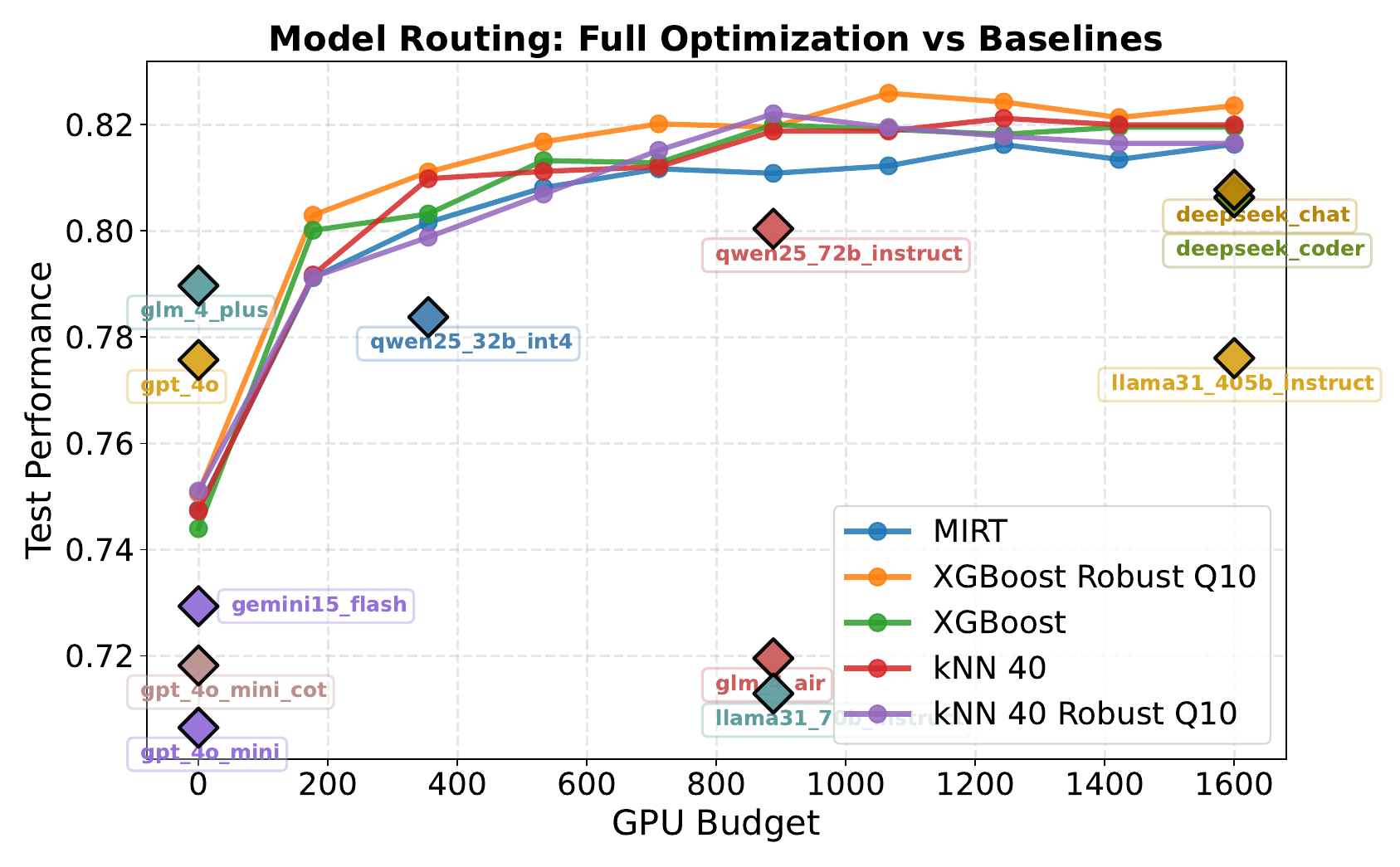}
    \caption{Dataset 1: $C=1\cdot 10^{-6}, B=200$}
\end{subfigure}
\hfill
\begin{subfigure}[b]{0.49\textwidth}
    \centering
    \includegraphics[width=\textwidth]{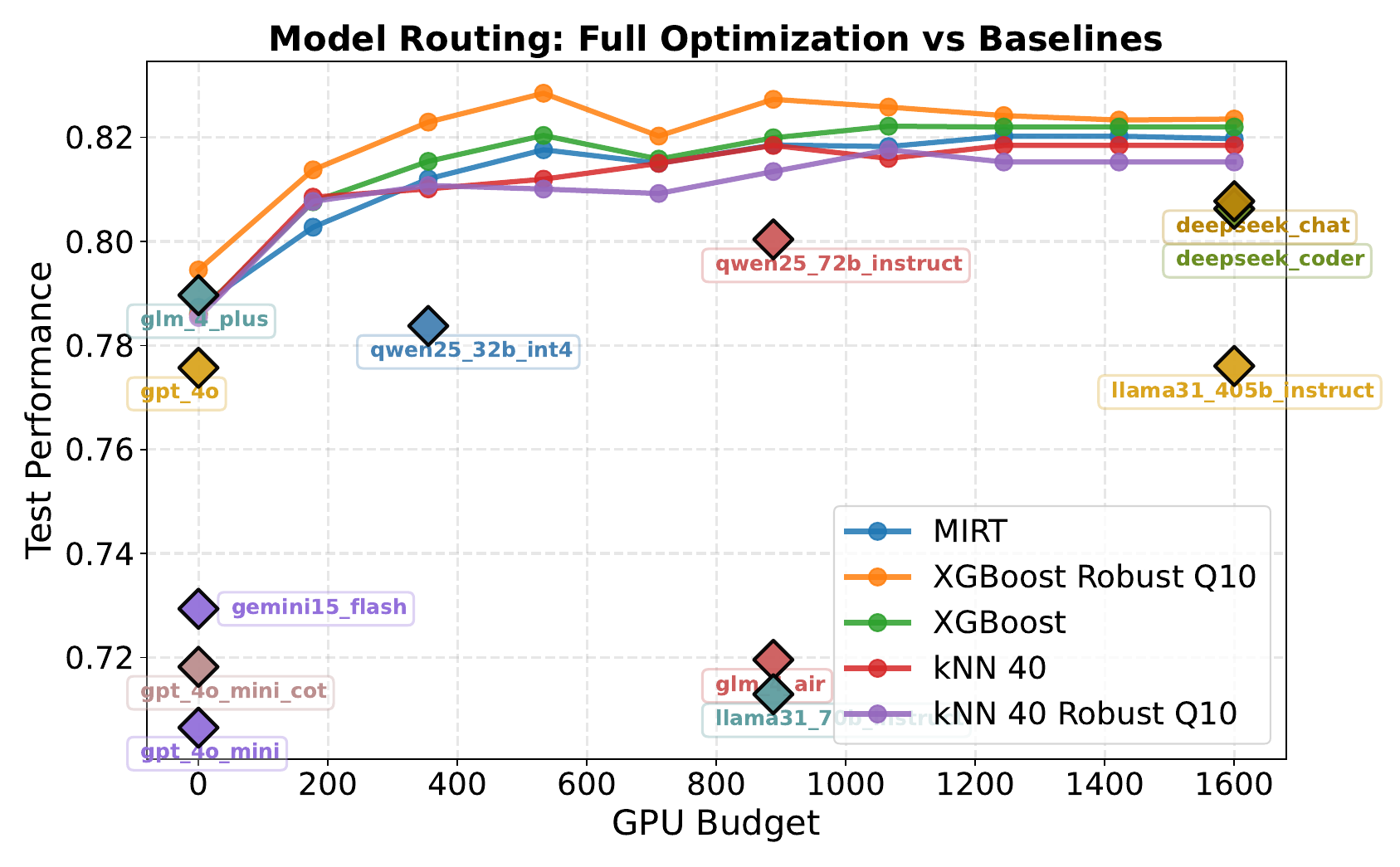}
    \caption{Dataset 1: $C=3\cdot 10^{-6}, B=200$}
\end{subfigure}

\vspace{0.5em}

\begin{subfigure}[b]{0.49\textwidth}
    \centering
    \includegraphics[width=\textwidth]{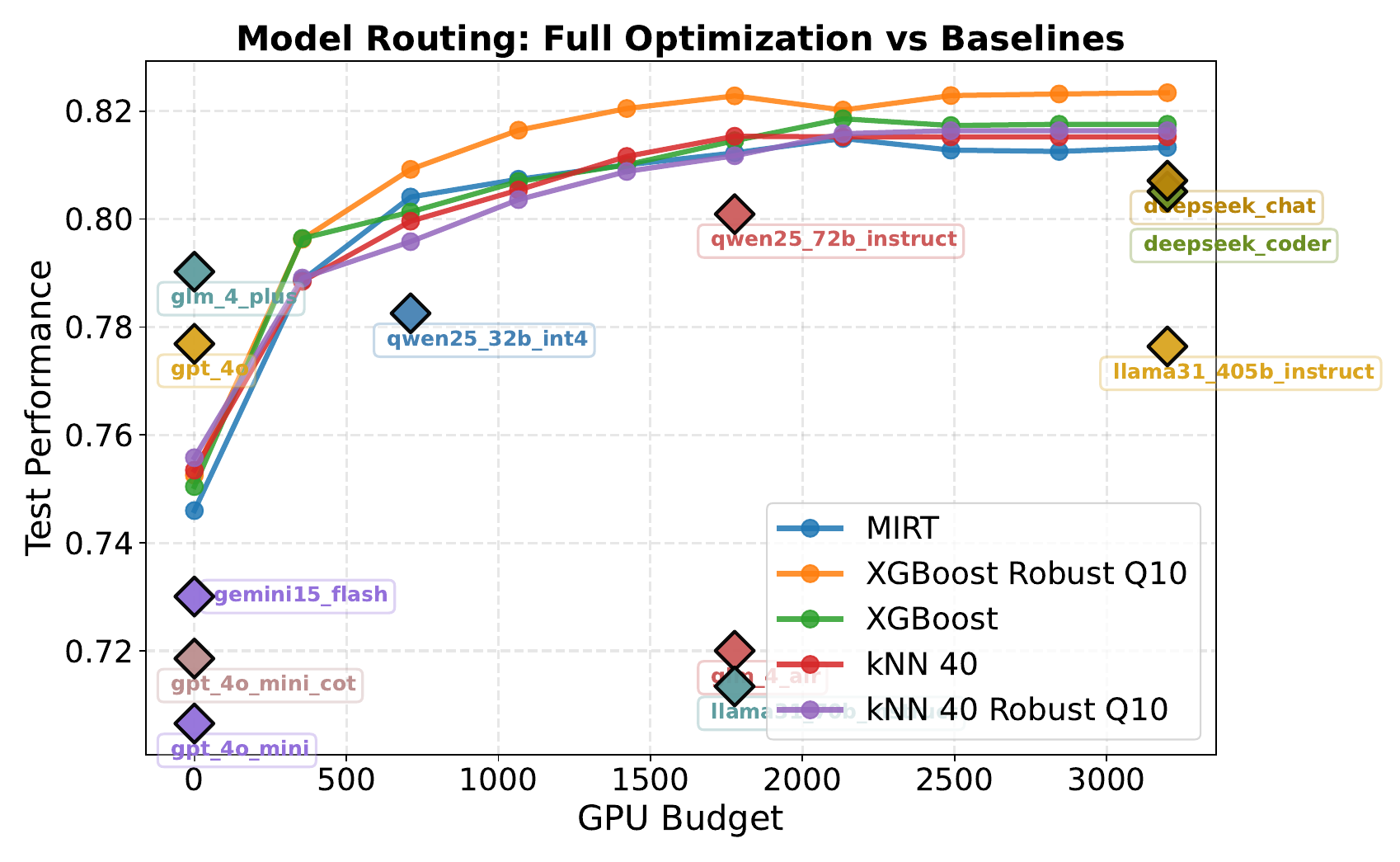}
    \caption{Dataset 1: $C=1\cdot 10^{-6}, B=400$}
\end{subfigure}
\hfill
\begin{subfigure}[b]{0.49\textwidth}
    \centering
    \includegraphics[width=\textwidth]{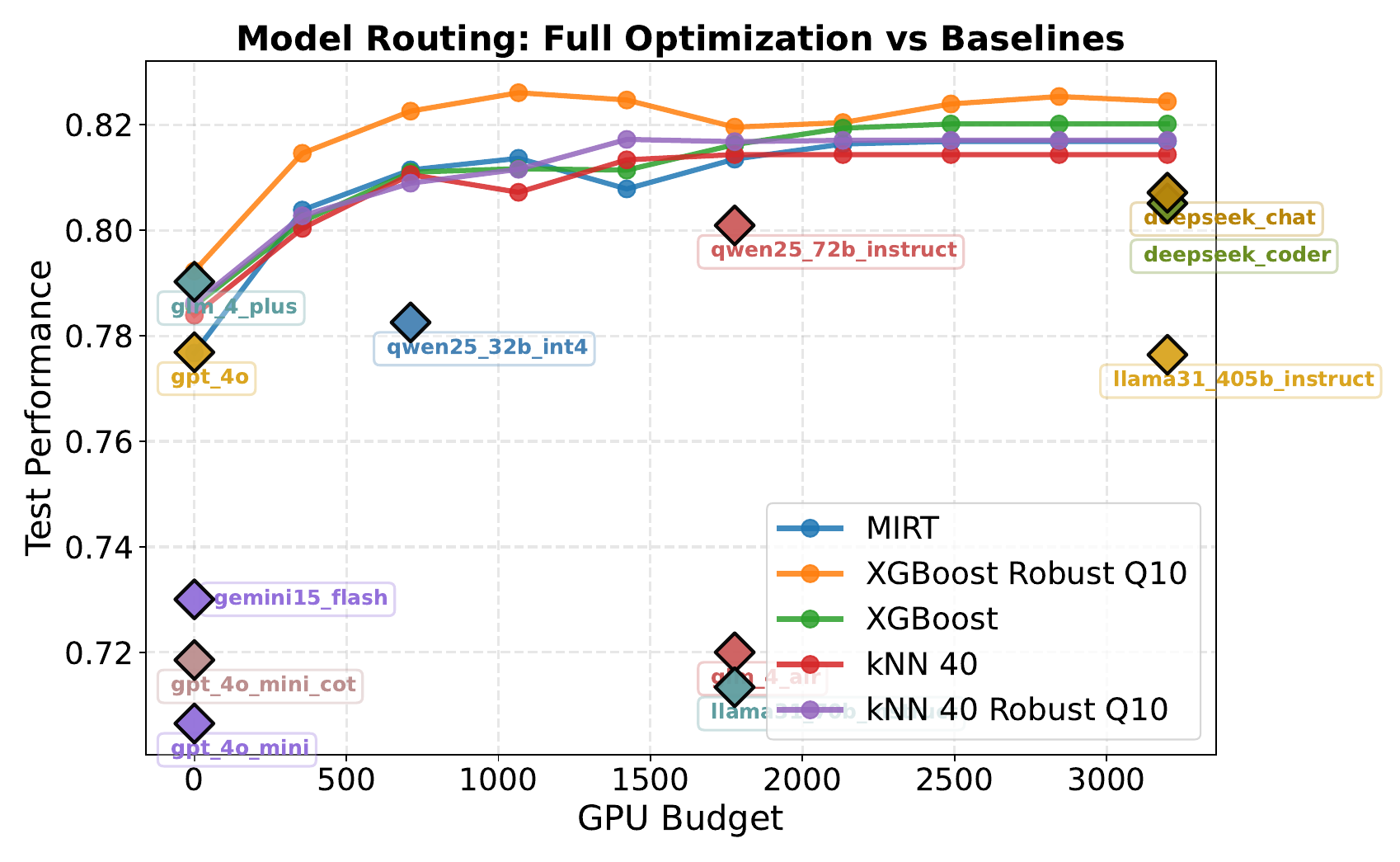}
    \caption{Dataset 1: $C=3\cdot 10^{-6}, B=400$}
\end{subfigure}

\caption{Full optimization results for Dataset 1 under varying batch sizes and cost budgets.}
\label{fig:full:optimization:dataset:1}
\end{figure}

\begin{figure}[htbp]
\centering

\begin{subfigure}[b]{0.49\textwidth}
    \centering
    \includegraphics[width=\textwidth]{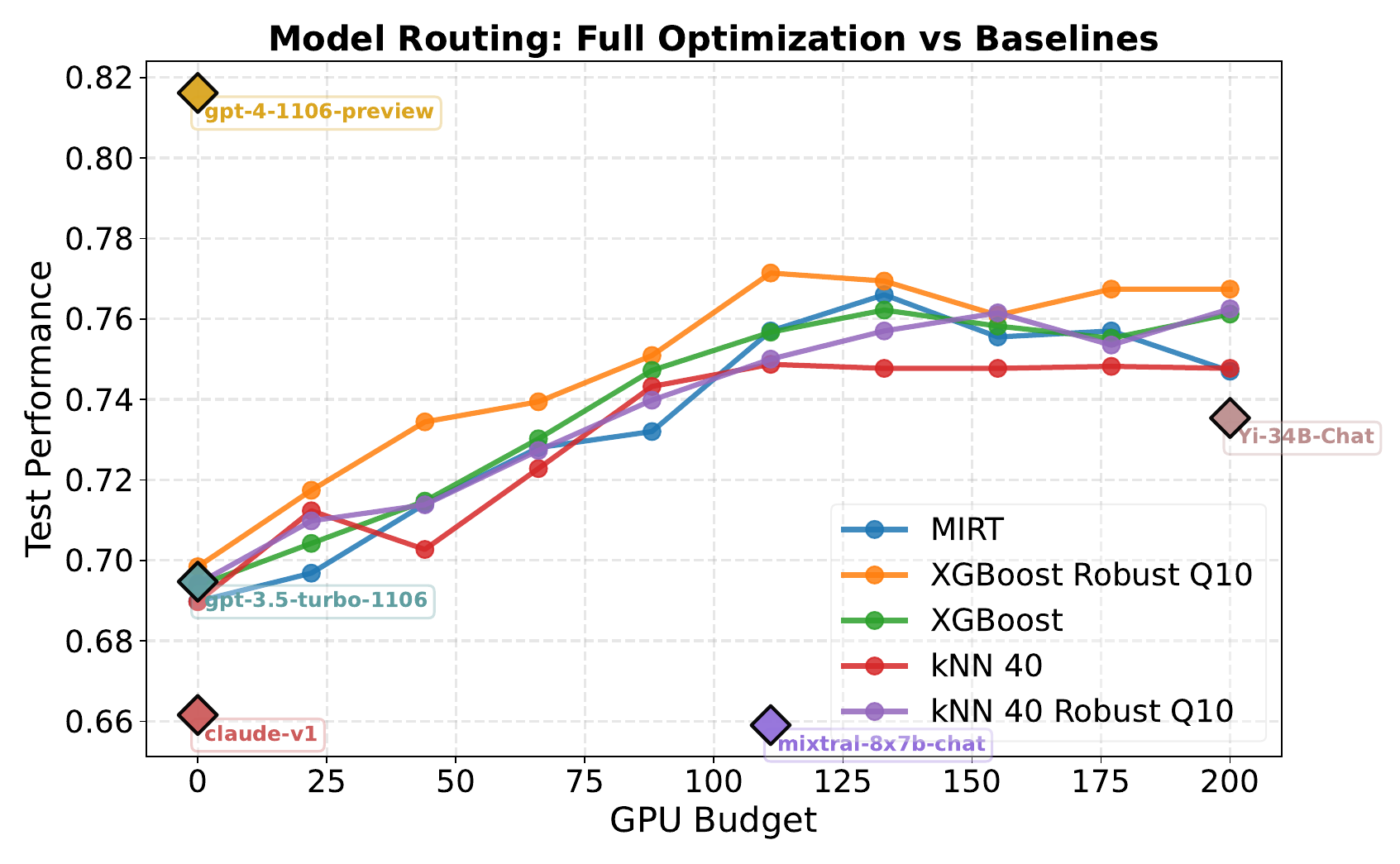}
    \caption{Dataset 2: $C=0.001, B=50$}
    \label{fig:full:optimization:dataset:2:B:50}
\end{subfigure}
\hfill
\begin{subfigure}[b]{0.49\textwidth}
    \centering
    \includegraphics[width=\textwidth]{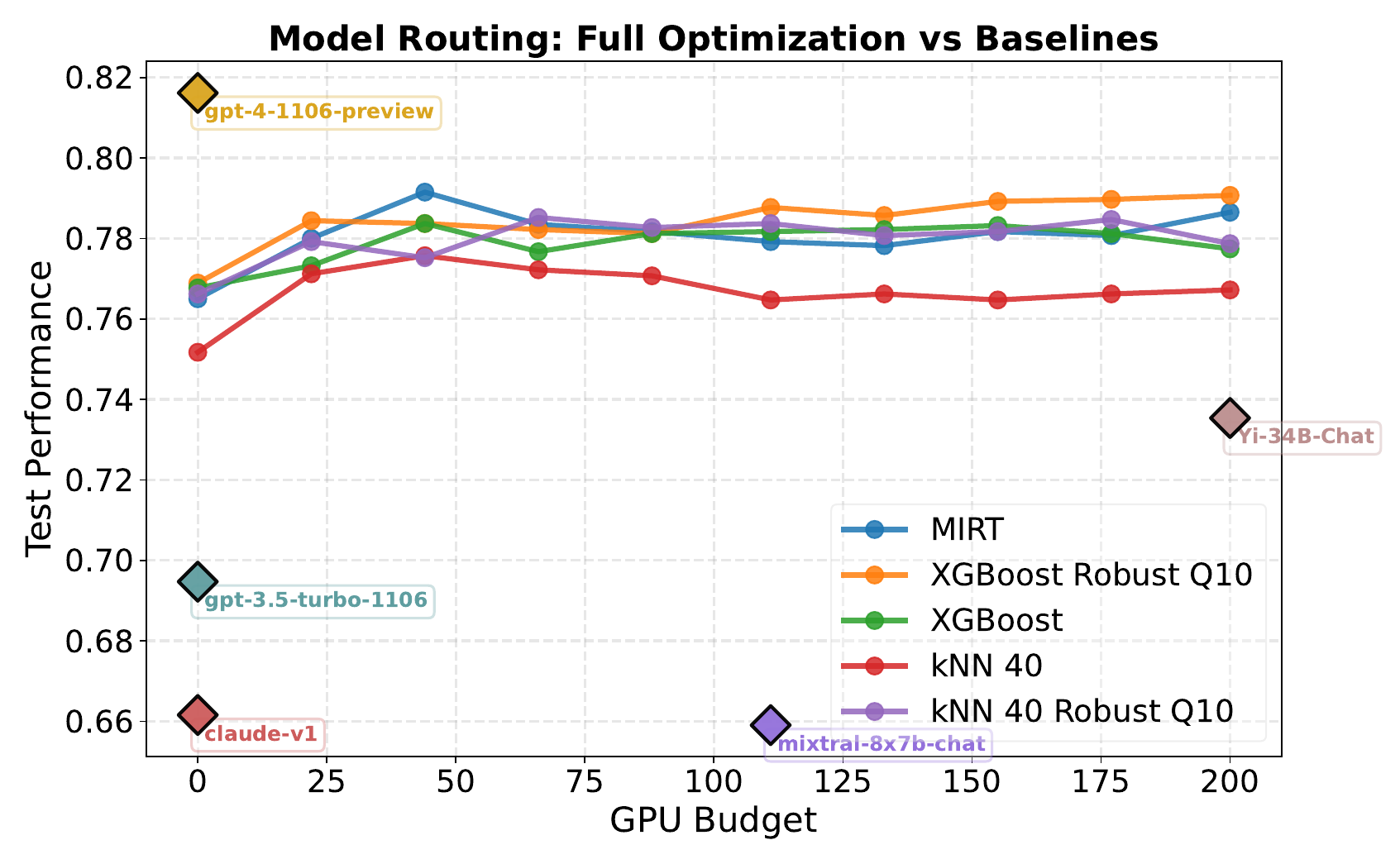}
    \caption{Dataset 2: $C=0.004, B=50$}
\end{subfigure}

\vspace{0.5em}

\begin{subfigure}[b]{0.49\textwidth}
    \centering
    \includegraphics[width=\textwidth]{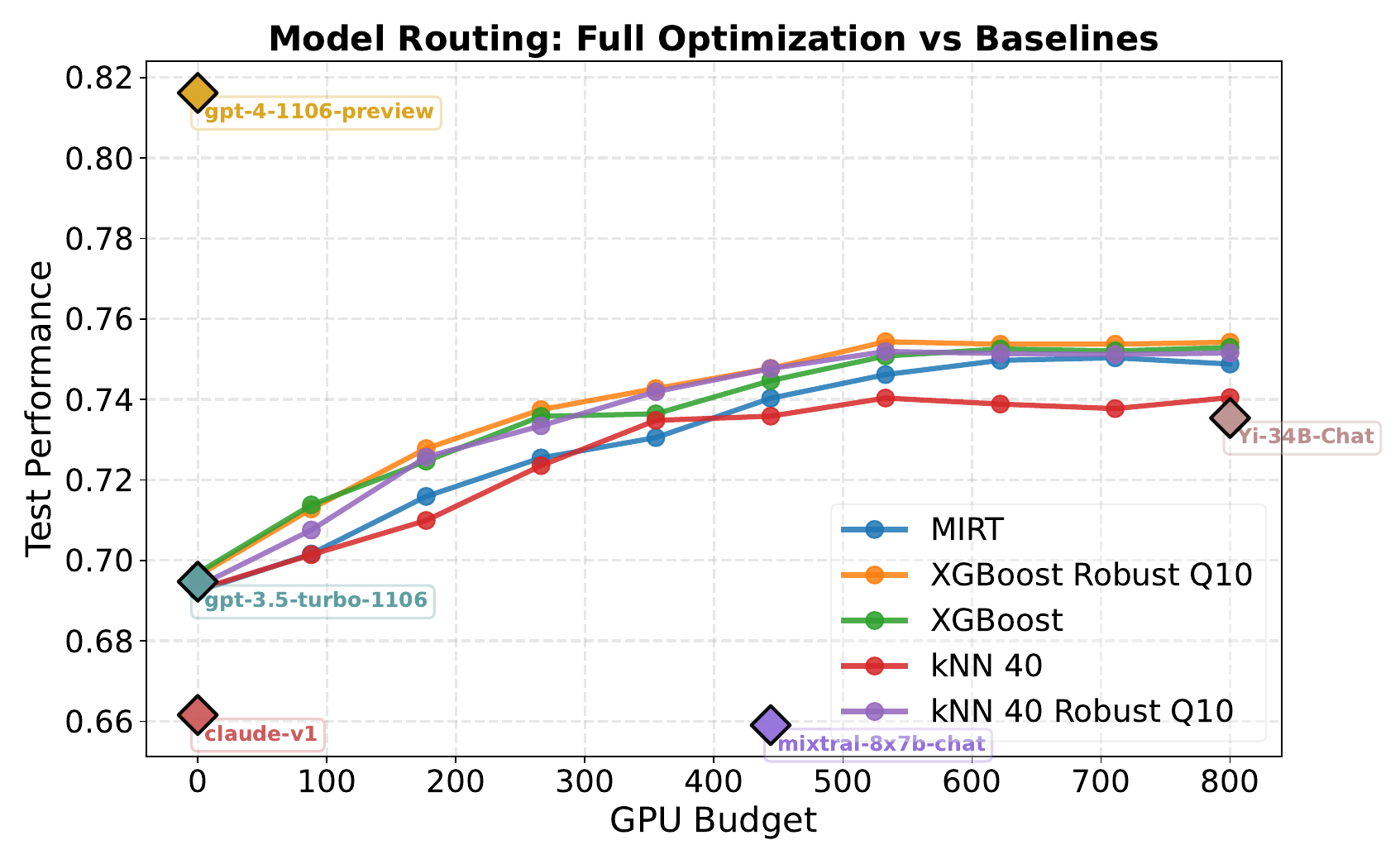}
    \caption{Dataset 2: $C=0.001, B=200$}
\end{subfigure}
\hfill
\begin{subfigure}[b]{0.49\textwidth}
    \centering
    \includegraphics[width=\textwidth]{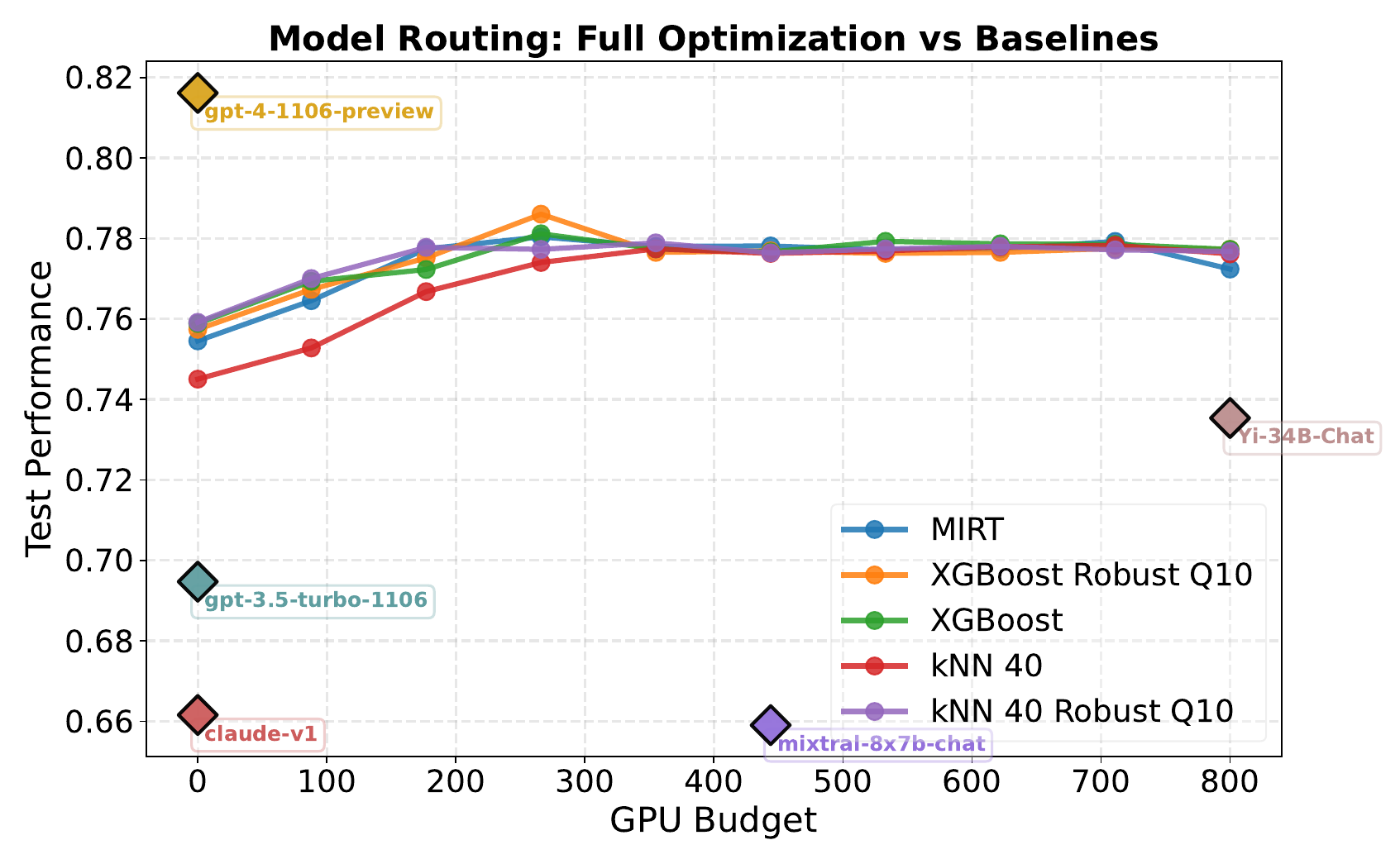}
    \caption{Dataset 2: $C=0.004, B=200$}
\end{subfigure}

\vspace{0.5em}

\begin{subfigure}[b]{0.49\textwidth}
    \centering
    \includegraphics[width=\textwidth]{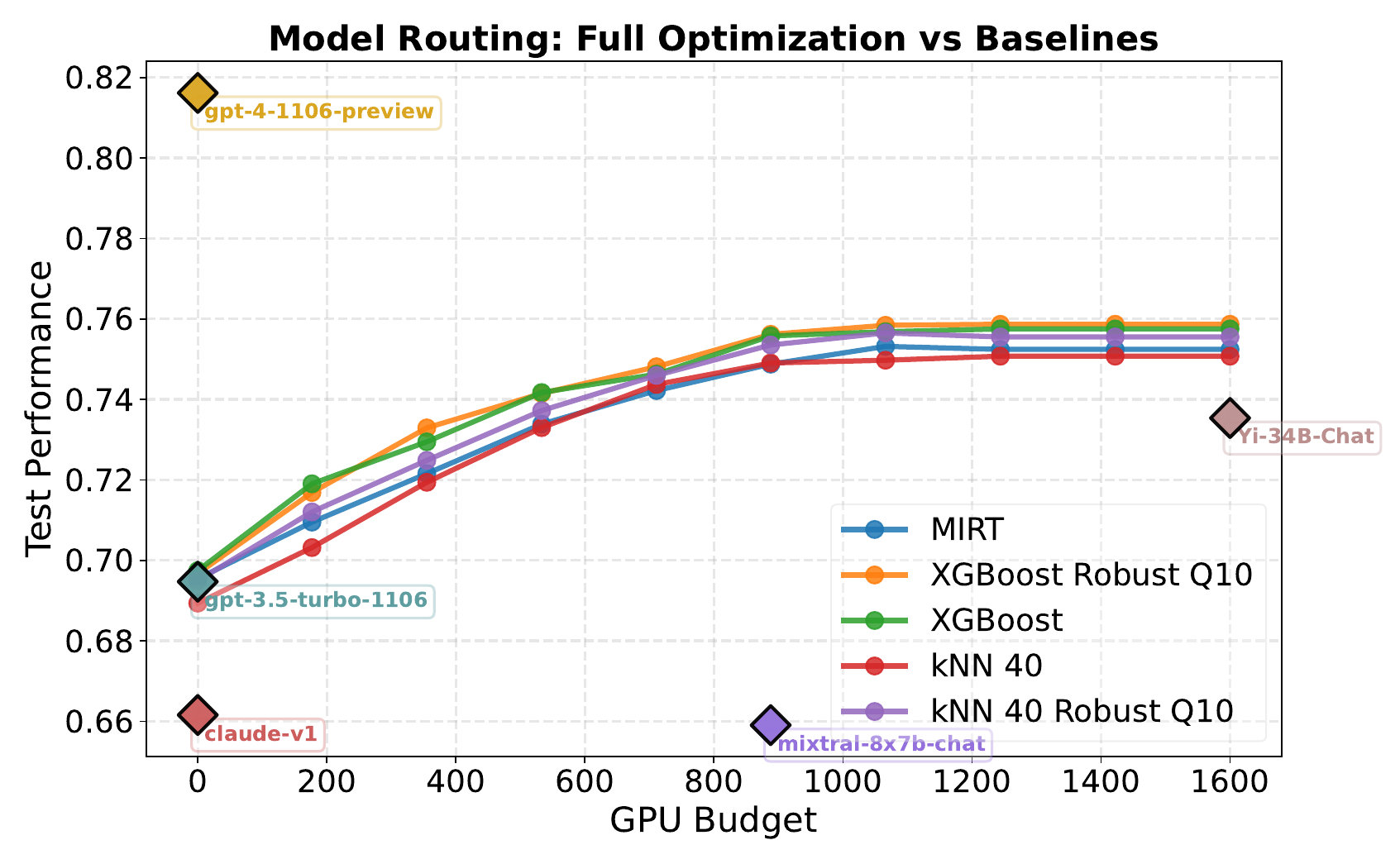}
    \caption{Dataset 2: $C=0.001, B=400$}
\end{subfigure}
\hfill
\begin{subfigure}[b]{0.49\textwidth}
    \centering
    \includegraphics[width=\textwidth]{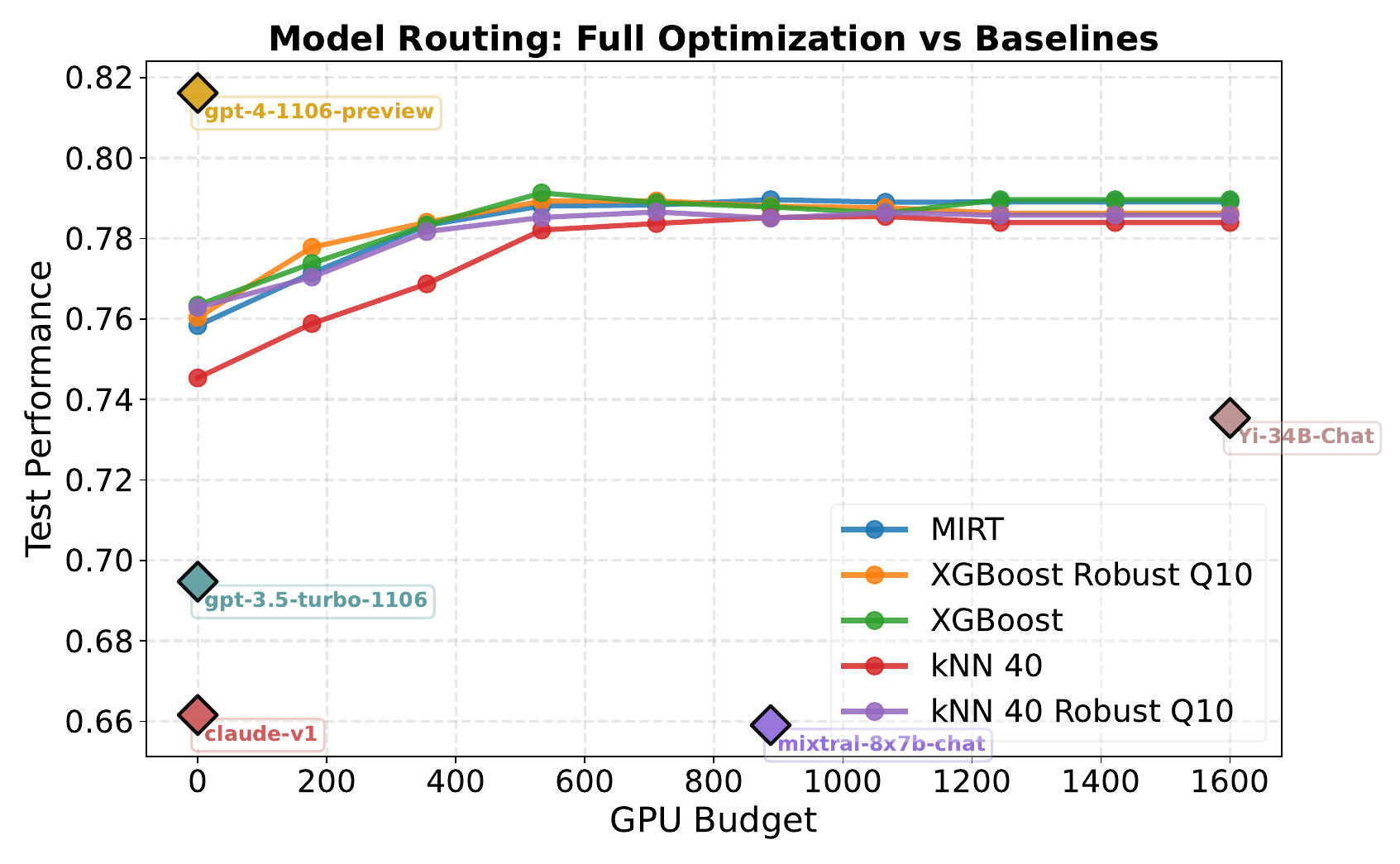}
    \caption{Dataset 2: $C=0.004, B=400$}
\end{subfigure}
\caption{Full optimization results for Dataset 2 under varying batch sizes and cost budgets. }
\label{fig:full:optimization:dataset:2}
\end{figure}

We present additional full-optimization experiments varying the batch size $B$ and the cost per query $C$. Results for $B \in \{50, 200, 400\}$ are shown in Figures~\ref{fig:full:optimization:dataset:1} and \ref{fig:full:optimization:dataset:2}, with low-cost and high-cost settings for each batch size.


\end{document}